\documentclass[journal,onecolumn,letterpaper]{IEEEtran} 

\usepackage[utf8]{inputenc} 
\usepackage[T1]{fontenc}
\usepackage{ifthen}
\usepackage{cite}
\usepackage[cmex10]{amsmath} 

\usepackage{times}  
\usepackage[hyphens]{url}  
\usepackage{algorithm}

\usepackage{newfloat}
\usepackage{listings}
\usepackage{subcaption}
\usepackage{afterpage}
\usepackage{makecell}
\usepackage{amsfonts}       
\usepackage{nicefrac}       

\usepackage[utf8]{inputenc} 

\usepackage[dvipsnames]{xcolor}
\RequirePackage[colorlinks,citecolor=RoyalBlue]{hyperref}

\usepackage{setspace}
\def\spacingset#1{\renewcommand{\baselinestretch}%
{#1}\small\normalsize}

\usepackage{algpseudocode}
\usepackage{paperpkg}

\newcommand{\calE}{\mathcal{E}}
\newcommand{\calF}{\mathcal{F}}
\newcommand{\calG}{\mathcal{G}}
\newcommand{\calH}{\mathcal{H}}
\newcommand{\calL}{\mathcal{L}}

\newcommand{\pushforward}{\#}

\DeclarePairedDelimiter{\norm}{\lVert}{\rVert}
\DeclarePairedDelimiter{\inner}{\langle}{\rangle}

\DeclareMathOperator{\PS}{\mathcal{P}}
\DeclareMathOperator{\diag}{diag}

\DeclareMathOperator{\softmax}{softmax}
\DeclareMathOperator{\Loss}{\mathcal{L}}
\DeclareMathOperator{\L2UVP}{\mathcal{L}^2-UVP}
\DeclareMathOperator{\CS}{CS}
\DeclareMathOperator{\id}{id}

\definecolor{OliveGreen}{RGB}{107, 142, 35}

\newcommand{\toprule}{\Xhline{2\arrayrulewidth}}
\newcommand{\midrule}{\hline}
\newcommand{\bottomrule}{\Xhline{2\arrayrulewidth}}

\usepackage{cleveref}

\begin{document}
\newcommand{\mytitle}{Embedding Empirical Distributions for Computing Optimal Transport Maps}

\title{\mytitle}


\author{

\IEEEauthorblockN{
Mingchen Jiang$^{2*}$\thanks{$^{*}$ Equal contribution.}
\qquad Peng Xu$^{3, 4*\dagger}$
\qquad Xichen Ye$^{1}$
\qquad Xiaohui Chen$^{4}$
\qquad Yun Yang$^{5}$
\qquad Yifan Chen$^{1\dagger}$
}

\IEEEauthorblockA{
$^{1}$ Hong Kong Baptist University \qquad
$^{2}$ Institute of Science Tokyo \qquad
$^{3}$ University of Illinois Urbana-Champaign \qquad \\
$^{4}$ University of Southern California \qquad
$^{5}$ University of Maryland, College Park \\
$^*$~~Equal contribution. 
This work was performed while the first author was interning at Hong Kong Baptist University. \\
$^\dagger$~~Correspondence to: 
Peng Xu \textlangle pengxu1@illinois.edu\textrangle,
Yifan Chen \textlangle yifanc@hkbu.edu.hk\textrangle.
}
}

\maketitle


\begin{abstract}
Distributional data have become increasingly prominent in modern signal processing, highlighting the necessity of computing optimal transport (OT) maps across \emph{multiple} probability distributions. 
Nevertheless, recent studies on neural OT methods predominantly focused on the efficient computation of a \emph{single} map between two distributions. 
To address this challenge, we introduce a novel approach to learning transport maps for new empirical distributions.
Specifically, we employ the transformer architecture to produce embeddings from distributional data of varying length;  
these embeddings are then fed into a hypernetwork to generate neural OT maps. 
Various numerical experiments were conducted to validate the embeddings and the generated OT maps. 
The model implementation and the code are provided on \url{https://github.com/jiangmingchen/HOTET}. 
\end{abstract}



\section{Introduction}\label{sec:intro}

Optimal transport (OT) theory~\cite{villani2003optimal} is a mathematical framework for finding the most efficient way (in the sense of minimizing a given cost function) to transport one probability distribution to another. When the quadratic cost is used, OT theory induces a metric space for probability measures, and the distance thereof is referred to as the 2-Wasserstein metric~\cite{panaretos2020invitation}. 
This notion provides a geometric view of distributions, and therefore makes OT an invaluable tool in information theory~\cite{7467523,9701924,10583905,10663506}. 
Furthermore, OT has already been used in many applications, such as flow-based diffusion models~\cite{lipman2023flow,liu2023flow}, GANs~\cite{salimans2018improving, xu2020cot}, style transfer~\cite{kolkin2019style}, data embedding~\cite{kolouri2021wasserstein,10.1093/imaiai/iaac023}, multilingual alignment~\cite{lian2021unsupervised, alqahtani2021using}, domain adaptation~\cite{courty2014domain, courty2017joint}, and model compression~\cite{singh2020model, wei2023ntk,resmoe}.



\begin{figure}[ht]
  \centering
  \includegraphics[width=0.7\columnwidth]{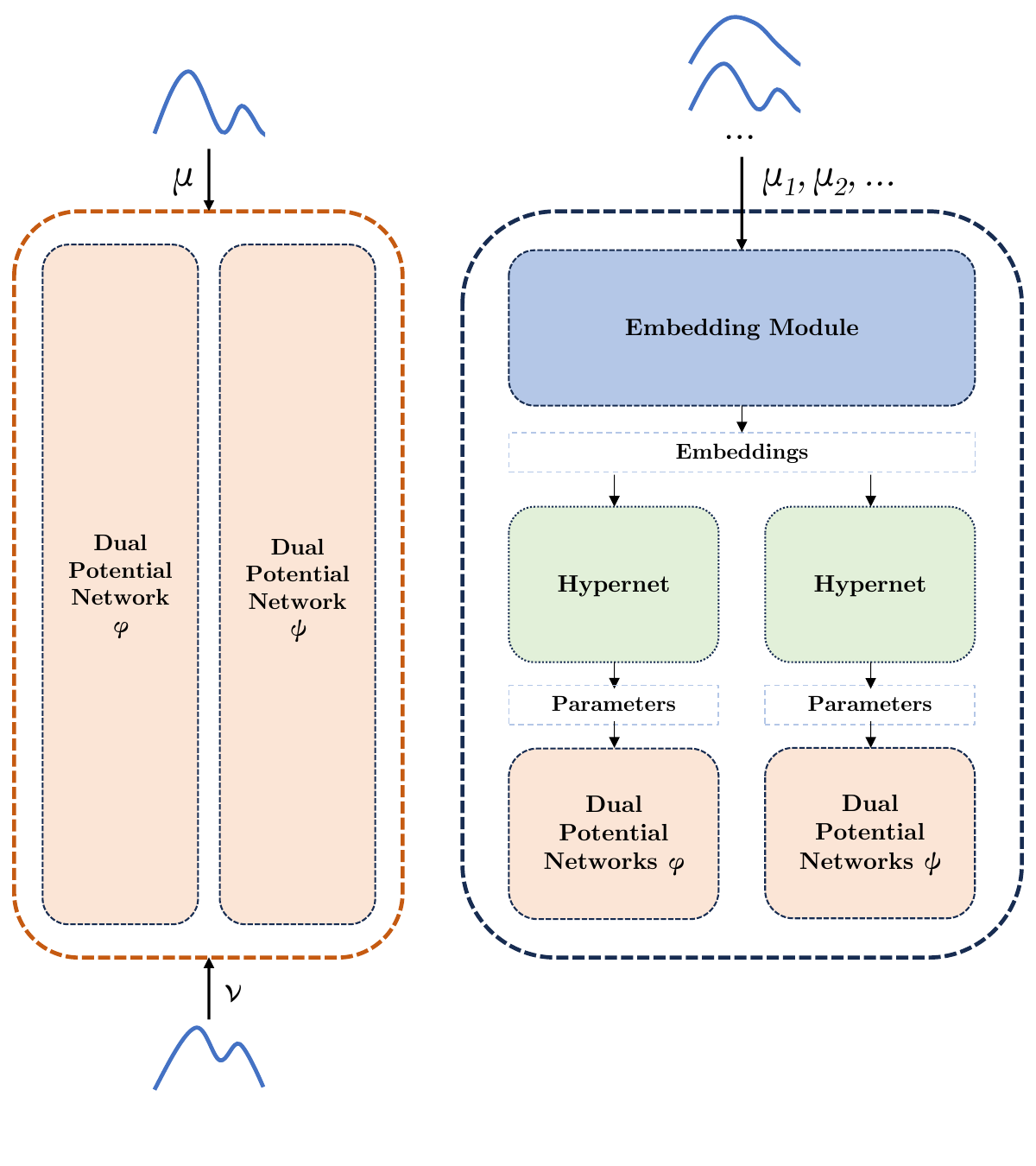}
  \caption{
  The network architectures for direct computation of OT maps (left), and for our proposed method of generating OT maps through hypernetworks (right).  \label{fig:model}}
  \vspace{-1.5em}
\end{figure}


\textbf{Challenges.}
In many applications, transport maps between \(n\) source distributions and a single target distribution are desired. 
For instance, in the Wasserstein embedding~\cite{kolouri2021wasserstein} scheme, input distributions are represented by the OT maps that link them to a reference distribution. 
Another example is the color transfer technique, where the color histogram of a reference image is transformed to match that of other images. 
To obtain OT maps in these settings, conventional approaches require the use of $2n$ neural networks to model the dual variables in the OT problem. 
If a new OT map is needed between a different source measure and the target measure, it must be computed from scratch. Computation of a single OT map is already challenging, and the difficulty increases considerably when seeking transport maps between numerous new source distributions and a single target distribution.

\textbf{Overview.}
To make the computation of multiple OT maps more efficient and generalizable, we propose a new paradigm (illustrated in the right panel of \Cref{fig:model}) to learn the OT maps between multiple source distributions \(\mu_1,\dotsc,\mu_n\) and a single target distribution \(\nu\).
In short, samples from the source distributions are passed to a transformer-based module \(\calE\) to produce embeddings in \(\bbR^{d}\). 
These embeddings are then feed into hypernetworks \(\calF\) and \(\calG\) to generate parameters for the potential networks, whose gradients approximate the OT maps. 
During training, samples from \(\nu\) are passed to the potential networks for loss computation, which propagates \(\nu\)'s information into \(\calF\) and \(\calG\). 
Notably, no explicit embedding of \(\nu\) is required after the training is complete.

We refer to the whole framework as the Hypernetworks for Optimal Transport with Embedding Transformers (\textbf{HOTET}). 
One notable strength of HOTET is that the embeddings of the input distributions in \(\bbR^d\) are directly obtained, 
which can then be used in downstream learning tasks.


\subsection{Related Works}


There are some works that use transformers to deal with OT problems, such as embedding empirical distributions into a latent space wherein Euclidean distances approximate OT distances\cite{haviv2024wasserstein}, regularizing the training process\cite{kan2025ot}, and neural network fusion\cite{imfeld2024transformerfusionoptimaltransport}. And our focus is on generating transport maps without engaging in direct computation by using transformers.      

Generally speaking, the idea of generating transport maps without direct computation is not new, as it is closely related to distributional regression \cite{doi:10.1080/01621459.2021.1956937,bonneel:hal-01303148}. Recent computational developments along this line include \textsc{CondOT} \cite{bunne2022supervised}, Meta OT~\cite{amos2022meta}, and GeONet \cite{gracyk2024geonet}.

In more detail, \textsc{CondOT}~\cite{bunne2022supervised} proposed to estimate a family of OT maps conditioned on a context variable, which can then be generalized given new context; 
in Meta OT~\cite{amos2022meta}, amortized optimization was used to predict OT maps from the input measure. Unlike the previous two, GeONet~\cite{gracyk2024geonet} learned neural operators to generate the \emph{Wasserstein geodesics} connecting the pair of input measures, which can be regarded as an extension to the aforementioned methods in the context of dynamic OT problem. However, this method falls out of the scope of neural OT learning, the focus of our project. 

We defer a more comprehensive comparison with existing methods to Appendix~\ref{app:comparison}.

\subsection{Our Contributions}

We summarize the contributions of this work as follows:
\begin{itemize}[leftmargin=*]
\item We proposed a new paradigm for learning neural OT maps and distribution embeddings for multiple distributions. 
\item We employed existing benchmarks~\cite{korotin2021neural} and conducted various tasks to demonstrate the effectiveness of our design.
\end{itemize}

\section{Preliminaries}

We start the review with the necessary notations for OT in \Cref{sec:prelim_ot}. 
We then introduce ICNN, hypernetwork, and Transformer, in Section~\ref{sec:prelim_icnn}. 
Moreover, a more comprehensive review, and the connection between attention and kernel estimators, are provided in Appendix~\ref{app:useful_facts};
building on these we propose to embed an empirical distribution (observation points with arbitrary sample weights) through a transformer in \Cref{sec:layer_embedding}, so as to generate embeddings for hypernetworks.
Notations in this work are collected in Appendix~\ref{sec:notation}.

\subsection{Optimal Transport}
\label{sec:prelim_ot}

Optimal transport, as the name suggests, is an optimization problem. 
We will review its properties in this subsection.

\textbf{Monge's problem.} 
Let \(\PS_2(\bbR^d) \) denote the set of probability measures over \(\bbR^d\) with finite second moments. Given \( \mu, \nu \in \PS_2(\bbR^d) \), Monge seeks a map \( T \) that push forwards \( \mu \) to \( \nu \) while 
minimizing the transportation cost. With cost \(c(x,y)=\norm{x-y}^2_2\), 
a metric $W_2$ on the space \(\PS_2(\Omega)\) is induced as
\begin{equation}\label{eqn:MP}
W_{2}^{2}(\mu, \nu)\coloneqq \inf_{T_\# \mu = \nu} \int_{\bbR^d} \norm{T(x) - x}_2^2 \,\dd\mu(x),\tag{MP}
\end{equation}
where \(T_{\#}\mu=\nu\) means that \(\nu(A) = \mu\bigl(T^{-1}(A)\bigr)\) for every Borel-measurable set \(A \subset \bbR^d\).

\textbf{Kantorovich's relaxation.} The constraint in~\eqref{eqn:MP} is highly nonlinear, which makes the solution difficult to obtain. Thus, a linear programming relaxation of \eqref{eqn:MP}, introduced by Kantorovich, is more commonly used in computation:
\begin{equation}\label{eqn:KR}
W_{2}^{2}(\mu, \nu)\coloneqq\inf_{\gamma \in \Pi(\mu,\nu)} \int_{\bbR^d \times \bbR^d}\norm{x - y}_2^2\,\dd\gamma(x, y). \tag{KR}
\end{equation}
The constraint set \(\Pi(\mu, \nu)\) in \eqref{eqn:KR} is the set of couplings between \(\mu\) and \(\nu\), i.e., probability measures \(\gamma \in \PS(\mathbb{R}^d \times \mathbb{R}^d)\) that satisfy
\((\pi_{x})_{\#}\gamma = \mu\) and \((\pi_{y})_{\#}\gamma = \nu\), where \(\pi_{x}(x, y) = x\), \(\pi_{y}(x, y) = y\) for all \((x, y) \in \mathbb{R}^d \times \mathbb{R}^d\). 

\textbf{Dual Formulation.} The relaxation \eqref{eqn:KR} admits a dual problem \cite[Thm.~1.3]{villani2003optimal}, which reads
\begin{equation}\label{eqn:DP}
\begin{aligned}
\frac{1}{2}W_{2}^{2}(\mu, \nu)=&\sup_{\varphi,\psi} &&\left\{ \int_{\bbR^d} \varphi\,\dd\mu+ \int_{\bbR^d} \psi \dd \nu \right\}&&\\
&\operatorname{\text{s.t.}} &&(\varphi,\psi) \in L_{\mu}^1 \times L_{\nu}^1&\\
& &&\varphi(x) + \psi(y) \leq \frac{1}{2}\norm{x - y}_2^2&
\end{aligned}\tag{DP}
\end{equation}
The optimal solution for the dual exists \cite[Prop.~1.11]{santambrogio2015} and is known as \emph{Kantorovich potentials}. 


The formulation \eqref{eqn:DP} and its extensions are adopted in multiple computational methods for their convenience. 
The conjugacy in the dual variables is typically enforced through regularization, cf.~\cite{korotin2021neural} for a comprehensive review. Once the optimal dual potentials $\varphi^\ast, \psi^\ast$ are obtained, the OT map (whenever exists) can be recovered as $T^\ast(x)=\nabla\Tilde{\varphi}(x)$, where $\Tilde{\varphi}(x)\coloneqq\norm{x}_2^2/2-\varphi^\ast(x)$ is a convex function. It follows that
\[(T^\ast)^{-1}(y)=y-\nabla\psi^\ast(y).\] In other words, the OT map and its inverse can be obtained by differentiating the solutions of \eqref{eqn:DP}.


\subsection{Neural Architectures}
\label{sec:prelim_icnn}

We then introduce the neural components in learning.

\textbf{Input Convex Neural Networks}~\cite[ICNN]{amos2017input} are utilized for modeling the convex potential function \(\Tilde{\varphi}(\cdot)\) in many neural OT implementations~\cite{korotin2021neural}. 
More details are deferred to Appendix~\ref{app:icnn}.

\textbf{Hypernetwork}~\cite{ha2017hypernetworks} is another component in our paradigm. 
Broadly speaking, hypernetwork refers to the neural network architecture designed to generate parameters for another neural network. Given a target network \(f_\theta\), instead of directly learning parameters \(\theta\) from data as traditional learning, 
hypernetworks generate \(\theta\) as an output mapped from a certain context variable. 
This allows for more flexible and efficient learning, particularly in tasks with complex or varying structures. Hypernetworks have been used in tasks such as neural architecture search, meta-learning, and conditional generation \cite{chauhan2023brief}.

\textbf{Transformers}~\cite{NIPS2017_3f5ee243}, equipped with the attention modules, are primarily used in natural language processing (NLP) tasks. 
The attention modules in transformers follow the spirit of nonparametric methods and can address sequences of indefinite length.
After removing the positional encoding for addressing sequences, transformers are proven universal approximators for set-to-set maps~\cite[Theorem~2]{Yun2020Are}, thus appropriate to handle empirical distributional data (a set of samples).
A complete introduction to transformers are deferred to Appendix~\ref{app:transformer_icnn}.


While the complexity of attention is quadratic, efficient GPU implementations, such as FlashAttention~\cite{dao2022flashattention, dao2023flashattention2}, have dramatically accelerated the computation while maintaining a linear space complexity. 
Our proposed method has incorporated the open-shelf efficient implementations as well.


\section{Hypernetworks for Optimal Transport with Embedding Transformers}

In this section, we discuss several crucial aspects of HOTET, our proposed training paradigm. Due to space limit, implementation details are deferred to Appendix~\ref{sec:imple-details}.





\subsection{Base OT solvers in HOTET}
\label{sec:training overview}


Training the hypernetworks in HOTET requires a base OT solver to learn to generate neural networks that approximate true OT maps 
(though in principle the HOTET framework is agnostic to the choice of the base solver, provided it delivers accurate approximations).
Mainstream OT solvers are mainly maximization-minimization-based, which are subject to divergence in training;
for the numerical experiments in this work, we selected the base solver in use from two instances for numerical stability, referred to as MM-B and MMv2 in \cite{korotin2021neural}. 
Both solvers utilize the dual OT formulation and ICNNs to approach the convex potentials \(\varphi\) and \(\psi\), whose gradients serve as approximations of the OT maps. 
Usage of the solvers is discussed in \Cref{subsec:exp_setups}, and detailed formulations / implementations for these solvers are provided in Appendix~\ref{app:base_solver}.


\subsection{Embedding empirical distributions}
\label{sec:layer_embedding}

The key step in our training paradigm is to generate context embeddings from the input \textbf{empirical distributions}. We employ transformers in this task for several reasons: 
\begin{enumerate}[leftmargin=*]
    \item With the positional encoding removed, transformers are universal approximators for set-to-set maps~\cite[Thm.~2]{Yun2020Are} and permutation invariant functions~\cite[Prop.~2]{lee2019set}, which match the characteristics of an empirical distribution.
    \item Transformers are suitable for inputs with variable sizes, which is not well handled by previous methods~\cite{amos2022meta,bunne2022supervised}. 
    \item The architecture of the transformer allows it to take the weights from the empirical distribution into consideration. This aspect is explored in Appendices~\ref{sec:attn-as-kernel} and~\ref{app:sample-weights}.
\end{enumerate}
A simplified process for HOTET to embed empirical distributions can be described as follows: a design matrix $\mtx X$ (the input empirical distribution) is passed to the $L$ blocks in our transformer embedding module; the first $L-1$ blocks follow a regular design in \cite{NIPS2017_3f5ee243}, which keeps the input sample dimension $hd$ ($h$ is the number of heads, see Appendix~\ref{app:transformer_icnn}); the MLP sub-layer in the last block lifts the hidden dimension to the one of the context vector. A mean-pooling of the output matrix produces the context vector, which is then fed into the hypernetworks to generate model parameters.


\subsection{HOTET for computing individual OT maps.}
\label{sec:hotet-individual}


To demonstrate the work flow of HOTET, we first discuss the computation of individual maps between two distributions with HOTET (though HOTET is \textbf{suboptimal} in this case).


HOTET consists of three modules: the embedding network \(\calE\) and two hypernetworks \(\calF,\calG\). Given a pair of empirical distributions $\mu, \nu$, the embedding network \(\calE\) is applied to generate context vectors $\calE(\mu), \calE(\nu)$ as specified in \Cref{sec:layer_embedding}. The two hypernetworks $\calF, \calG$ will then respectively take the context $\calE(\mu), \calE(\nu)$ as inputs, and produce parameters \[\theta = \calF\circ\calE(\mu), \quad 
\omega = \calG\circ\calE(\nu),\] for the two ICNNs $f_\theta(\cdot)$ and $g_\omega(\cdot)$ approximating the convex potentials. 
Afterwards, the potential networks are provided to the base OT solver, which concludes the entire forward pass. 
The resulting gradients of the forward pass are then used to update the three modules (\(\calE\) and $\calF, \calG$) with backpropagation.

As the parameter space for $\theta, \omega$ in HOTET is strictly smaller than in the original solver, there will be natural concerns about the representation power of HOTET.
Shortly in \Cref{subsec:W2B}, we compare HOTET with the MMv2 solver as a sanity check, under the same setting taken by~\cite{korotin2021neural};
we note the proposed framework still achieves a comparable performance in unfavorable settings.
The details are provided in Appendix~\ref{app:training_algo}.

\begin{table}[!t]
    \centering
    \caption{\label{tab:uvp_w2b}
    Performance of the constructed forward (fwd) and inverse (inv) transport maps ($\calL^2$-UVP (\%) as the metric). 
    Lower implies the fitted map approximates the true OT map better. 
    The standard deviation is calculated over 10 runs.
    }
    \scalebox{0.90}
    {
    \begin{tabular}{c|ccc|ccc}
        \toprule
        \thead{DIM}  & \thead{HOTET (fwd)} & \thead{MetaOT (fwd)} & \thead{MM-B (fwd)} & \thead{HOTET (inv)} & \thead{MetaOT (inv)}  & \thead{MM-B (inv)}\\
\midrule
2 & 5.03 {\textcolor{gray}{\tiny ±0.33}} & 11.71 {\textcolor{gray}{\tiny ±0.49}} & 0.54 {\textcolor{gray}{\tiny ±0.06}} & 0.66 {\textcolor{gray}{\tiny ±0.03}} & 15.52 {\textcolor{gray}{\tiny ±0.56}} & 0.81 {\textcolor{gray}{\tiny ±0.03}} \\
4 & 10.06 {\textcolor{gray}{\tiny ±0.39}} & 11.63 {\textcolor{gray}{\tiny ±0.46}} & 1.91 {\textcolor{gray}{\tiny ±0.10}} & 2.01 {\textcolor{gray}{\tiny ±0.04}} & 9.27 {\textcolor{gray}{\tiny ±0.42}} & 0.48 {\textcolor{gray}{\tiny ±0.07}} \\
8 & 11.38 {\textcolor{gray}{\tiny ±0.41}} & 23.18 {\textcolor{gray}{\tiny ±0.42}} & 5.51 {\textcolor{gray}{\tiny ±0.15}} & 3.80 {\textcolor{gray}{\tiny ±0.23}} & 20.50 {\textcolor{gray}{\tiny ±0.87}} & 2.22 {\textcolor{gray}{\tiny ±0.05}} \\
16 & 16.91 {\textcolor{gray}{\tiny ±0.58}} & 47.61 {\textcolor{gray}{\tiny ±1.03}} & 13.17 {\textcolor{gray}{\tiny ±0.08}} & 4.64 {\textcolor{gray}{\tiny ±0.18}} & 34.16 {\textcolor{gray}{\tiny ±0.47}} & 6.27 {\textcolor{gray}{\tiny ±0.32}} \\
32 & 20.87 {\textcolor{gray}{\tiny ±0.56}} & 59.99 {\textcolor{gray}{\tiny ±0.67}} & 25.87 {\textcolor{gray}{\tiny ±0.59}} & 17.03 {\textcolor{gray}{\tiny ±0.47}} & 44.60 {\textcolor{gray}{\tiny ±0.78}} & 10.20 {\textcolor{gray}{\tiny ±0.26}} \\
64 & 37.62 {\textcolor{gray}{\tiny ±0.84}} & 74.02 {\textcolor{gray}{\tiny ±0.80}} & 23.63 {\textcolor{gray}{\tiny ±0.25}} & 15.07 {\textcolor{gray}{\tiny ±0.37}} & 30.88 {\textcolor{gray}{\tiny ±0.89}} & 13.95 {\textcolor{gray}{\tiny ±0.52}} \\
\bottomrule
    \end{tabular}
    }
\end{table}

\begin{table}[!t]
    \centering
    \caption{\label{tab:cos_w2b}
    Cosine similarity between $\hat{T}$ and $T^*$. 
    The standard deviation is calculated over 10 runs.
    Best viewed zoom in. 
    }
    \scalebox{0.90}{
    \begin{tabular}{c|ccc|ccc}
        \toprule
        \thead{DIM}  & \thead{HOTET (fwd)} & \thead{MetaOT (fwd)} & \thead{MM-B (fwd)} & \thead{HOTET (inv)} & \thead{MetaOT (inv)}  & \thead{MM-B (inv)}\\
        \midrule  
2 & 0.93 {\textcolor{gray}{\tiny ±0.01}} & 0.81 {\textcolor{gray}{\tiny ±0.01}} & 0.99 {\textcolor{gray}{\tiny ±0.01}}  & 0.97 {\textcolor{gray}{\tiny ±0.01}} & 0.74 {\textcolor{gray}{\tiny ±0.01}} & 0.99 {\textcolor{gray}{\tiny ±0.01}} \\
4 & 0.87 {\textcolor{gray}{\tiny ±0.01}} & 0.84 {\textcolor{gray}{\tiny ±0.01}} & 0.96 {\textcolor{gray}{\tiny ±0.01}} & 0.96 {\textcolor{gray}{\tiny ±0.01}}& 0.88 {\textcolor{gray}{\tiny ±0.01}} & 0.97 {\textcolor{gray}{\tiny ±0.01}} \\
8 & 0.89 {\textcolor{gray}{\tiny ±0.01}} & 0.80 {\textcolor{gray}{\tiny ±0.01}} & 0.94 {\textcolor{gray}{\tiny ±0.01}}& 0.97{\textcolor{gray}{\tiny ±0.01}} & 0.82 {\textcolor{gray}{\tiny ±0.01}} & 0.97 {\textcolor{gray}{\tiny ±0.01}} \\
16 & 0.90 {\textcolor{gray}{\tiny ±0.01}} & 0.74 {\textcolor{gray}{\tiny ±0.01}} & 0.92 {\textcolor{gray}{\tiny ±0.01}} & 0.97 {\textcolor{gray}{\tiny ±0.01}}& 0.78 {\textcolor{gray}{\tiny ±0.01}} & 0.96 {\textcolor{gray}{\tiny ±0.01}} \\
32 & 0.91 {\textcolor{gray}{\tiny ±0.01}} & 0.76 {\textcolor{gray}{\tiny ±0.01}} & 0.90 {\textcolor{gray}{\tiny ±0.01}} & 0.91 {\textcolor{gray}{\tiny ±0.01}} & 0.80 {\textcolor{gray}{\tiny ±0.01}} & 0.97 {\textcolor{gray}{\tiny ±0.01}} \\
64 & 0.84 {\textcolor{gray}{\tiny ±0.01}} & 0.71 {\textcolor{gray}{\tiny ±0.01}} & 0.90 {\textcolor{gray}{\tiny ±0.01}} & 0.91 {\textcolor{gray}{\tiny ±0.01}} & 0.87 {\textcolor{gray}{\tiny ±0.01}} & 0.94 {\textcolor{gray}{\tiny ±0.01}} \\
        \noalign{\hrule height 0.8pt}  
    \end{tabular}
    }
\end{table}

\subsection{HOTET for computing multiple OT maps}
\label{sec:hotet-multi}



We then illustrate the desired scenario for HOTET where the source measures \(\mu_1,\dotsc,\mu_n\) (training set) and the reference measure \(\nu\) are already given and one aims to efficiently obtain the corresponding OT maps between a new $\mu$ and the reference~$\nu$.
In training, instead of computing OT maps individually for the $n$ distribution pairs, HOTET generates the \(2n\) sets of parameters for ICNNs together via \(\calF\) and \(\calG\). To train the hypernetworks, loss from the base OT solver is computed for each \((\mu_i,\nu)\) and subsequently aggregated together.





Since the reference measure $\nu$ is fixed in the training and testing process, we take $\mu_i$ as input to model \textbf{both the forward map and the inverse map} between $\mu_i$ and $\nu$. That is, the parameters for the ICNNs $f_{\theta_i}$ (resp.~$g_{\omega_i}$) are generated as $\theta_i=\calF\circ\calE({\mu}_i)$ (resp.~$\omega_i=\calG\circ\calE({\mu}_i)$)

\begin{table*}[!t]
    \centering
    \small
    \caption{Performance of the constructed transport forward (fwd) and inverse (inv) maps in predicting OT maps ($\calL^2$-UVP~(\%) as the metric). 
    The standard deviation is calculated over 10 runs.
    \label{tab:uvp_mix_gaussians}
    }
    \resizebox{0.9\textwidth}{!}{
    \begin{tabular}{c|cc|cc|cc|cc}
        \toprule
        \multirow{2}{*}{\thead{DIM}} & \multicolumn{4}{c|}{\thead{Train}} & \multicolumn{4}{c}{\thead{Predict}} \\
        \cline{2-9}
          & \thead{HOTET (fwd)} & \thead{MetaOT (fwd)} & \thead{HOTET (inv)} & \thead{MetaOT (inv)} & \thead{HOTET (fwd)} & \thead{MetaOT (fwd)} & \thead{HOTET (inv)} & \thead{MetaOT (inv)} \\
\midrule
2 & 3.25 {\textcolor{gray}{\tiny ±0.56}} & 2.66 {\textcolor{gray}{\tiny ±0.41}} & 3.01 {\textcolor{gray}{\tiny ±0.49}} & 2.71 {\textcolor{gray}{\tiny ±0.43}} & 3.13 {\textcolor{gray}{\tiny ±0.47}} & 2.63 {\textcolor{gray}{\tiny ±0.45}} & 3.07 {\textcolor{gray}{\tiny ±0.45}} & 2.69 {\textcolor{gray}{\tiny ±0.47}} \\
4 & 3.40 {\textcolor{gray}{\tiny ±0.29}} & 20.27 {\textcolor{gray}{\tiny ±2.12}} & 3.44 {\textcolor{gray}{\tiny ±0.30}} & 23.37 {\textcolor{gray}{\tiny ±2.43}} & 3.37 {\textcolor{gray}{\tiny ±0.27}} & 20.27 {\textcolor{gray}{\tiny ±2.11}} & 3.43 {\textcolor{gray}{\tiny ±0.29}} & 23.48 {\textcolor{gray}{\tiny ±2.46}} \\
8 & 6.59 {\textcolor{gray}{\tiny ±0.28}} & 50.34 {\textcolor{gray}{\tiny ±0.81}} & 6.48 {\textcolor{gray}{\tiny ±0.27}} & 54.26 {\textcolor{gray}{\tiny ±1.45}} & 6.54 {\textcolor{gray}{\tiny ±0.31}} & 50.35 {\textcolor{gray}{\tiny ±0.82}} & 6.45 {\textcolor{gray}{\tiny ±0.28}} & 54.29 {\textcolor{gray}{\tiny ±1.48}} \\
16 & 10.72 {\textcolor{gray}{\tiny ±0.26}} & 73.76 {\textcolor{gray}{\tiny ±0.82}} & 11.19 {\textcolor{gray}{\tiny ±0.27}} & 74.02 {\textcolor{gray}{\tiny ±1.15}} & 10.59 {\textcolor{gray}{\tiny ±0.25}} & 73.70 {\textcolor{gray}{\tiny ±0.82}} & 11.05 {\textcolor{gray}{\tiny ±0.25}} & 73.98 {\textcolor{gray}{\tiny ±1.16}} \\
32 & 18.00 {\textcolor{gray}{\tiny ±0.47}} & 86.93 {\textcolor{gray}{\tiny ±0.37}} & 18.96 {\textcolor{gray}{\tiny ±0.54}} & 88.27 {\textcolor{gray}{\tiny ±0.35}} & 18.03 {\textcolor{gray}{\tiny ±0.44}} & 86.92 {\textcolor{gray}{\tiny ±0.37}} & 19.00 {\textcolor{gray}{\tiny ±0.53}} & 88.27 {\textcolor{gray}{\tiny ±0.33}} \\
64 & 29.18 {\textcolor{gray}{\tiny ±0.46}} & 93.13 {\textcolor{gray}{\tiny ±0.25}} & 26.44 {\textcolor{gray}{\tiny ±0.36}} & 94.13 {\textcolor{gray}{\tiny ±0.40}} & 28.29 {\textcolor{gray}{\tiny ±0.83}} & 93.14 {\textcolor{gray}{\tiny ±0.25}} & 26.49 {\textcolor{gray}{\tiny ±0.35}} & 94.11 {\textcolor{gray}{\tiny ±0.41}} \\
\bottomrule
    \end{tabular}
}
\end{table*}

\section{Numerical Experiments}
\label{sec:Evaluation}

We conducted various experiments to evaluate the performance of HOTET. The detailed experimental setups are discussed in \Cref{subsec:exp_setups}. In particular, we perform a sanity check (\Cref{subsec:W2B}) based on the benchmarks from~\cite{korotin2021neural} to exhibit the capability of our paradigm to generate quality transport map. 
This particular setting will be referred to as \textbf{W2B} hereinafter. Furthermore, we evaluated the prediction performance of our proposed training paradigm in \Cref{subsec:MGD}, where we demonstrated that HOTET is capable of generating transport maps for unseen distributions after being trained on similar sample distributions (c.f.~\Cref{sec:hotet-multi}). Lastly, we tested our model on images data through various applications in \Cref{subsec:color_transfer}, and performed an ablation study on the embedding module in \Cref{subsec:ablation}.

\subsection{Experiment Settings}
\label{subsec:exp_setups}



\subsubsection{Choice of base OT solvers}
While the theory foundation for OT is solid, the choice of a better solver for a specific task is an engineering problem. It is typical that in some cases the solvers will be hard to optimize: 
for example, the MMv2 solver tends to produce forward and inverse maps with large performance discrepancy, due to its asymmetric nature (detailed in Appendix~\ref{app:base_solver}).
When applicable, we compare the performance of the MM-B and MMv2 solver (see Appendix~\ref{app:solvers}), and report the results of the more stable one in the main text.





\subsubsection{W2B benchmark} The performance of numerous OT solvers are evaluated in \cite{korotin2021neural} on Gaussian mixture distributions. 
The performance of an estimated transport map \(\hat{T}\) with respect to the ground truth \(T^\ast\) is evaluated with the \(\calL^2\)-Unexplained Variance Percentage (UVP) and Cosine Similarity (CS): 
\begin{align*}
& \L2UVP(\hat{T})\coloneqq100\cdot\frac{\norm{\hat{T} - T^*}^2_{\calL^2(\mu)}}{\operatorname{Var}(\nu)}\%, \\
& \CS(\hat{T})\coloneqq\frac{\inner{\hat{T}-\id,\nabla\psi^\ast-\id}_{\calL^2(\mu)}}{\norm{\hat{T} - \id}^2_{\calL^2(\mu)}\cdot\norm{T^*-\id}^2_{\calL^2(\mu)}}.
\end{align*}
We inherited these settings for our own comparisons.

\subsubsection{Predicting OT maps with HOTET} In this experiment, multiple Gaussian mixture distributions are generated with random means and covariance matrices. One distribution is designated as the reference, while the rest are divided into training and test sets. Distributions in the test set are used to assess the quality of the OT maps predicted by HOTET.


\subsubsection{Color transfer} To further explore the capabilities of HOTET, we conducted color transfer experiments on paintings from \href{https://www.wikiart.org/}{WikiArt}. The experiments include transferring colors from one image to another, as well as transferring colors from multiple images to a single target image.

\begin{figure}[t]
    \centering
    \begin{subfigure}{0.40\columnwidth}
        \includegraphics[width=\textwidth]{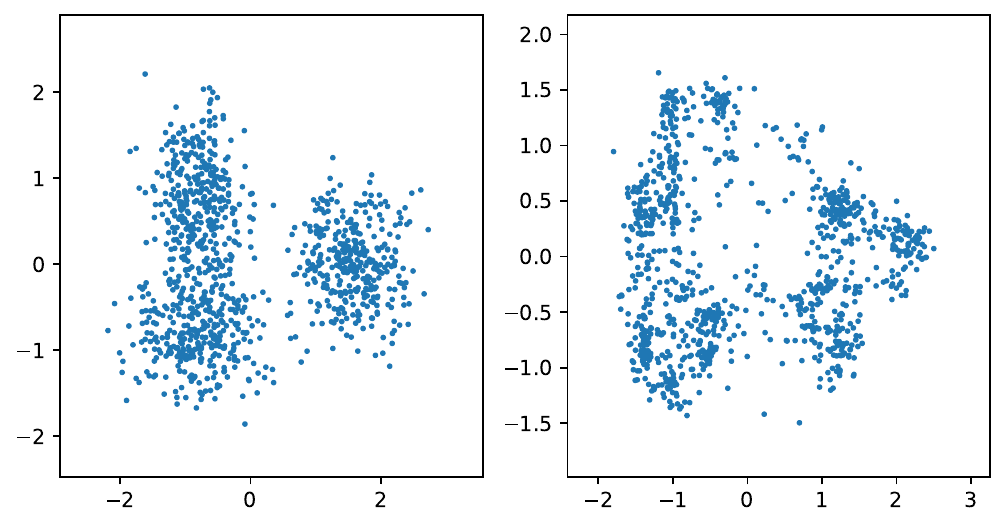}
        \caption{Ground Truth}
    \end{subfigure}
    \hfill
    \begin{subfigure}{0.40\columnwidth}
        \includegraphics[width=\textwidth]{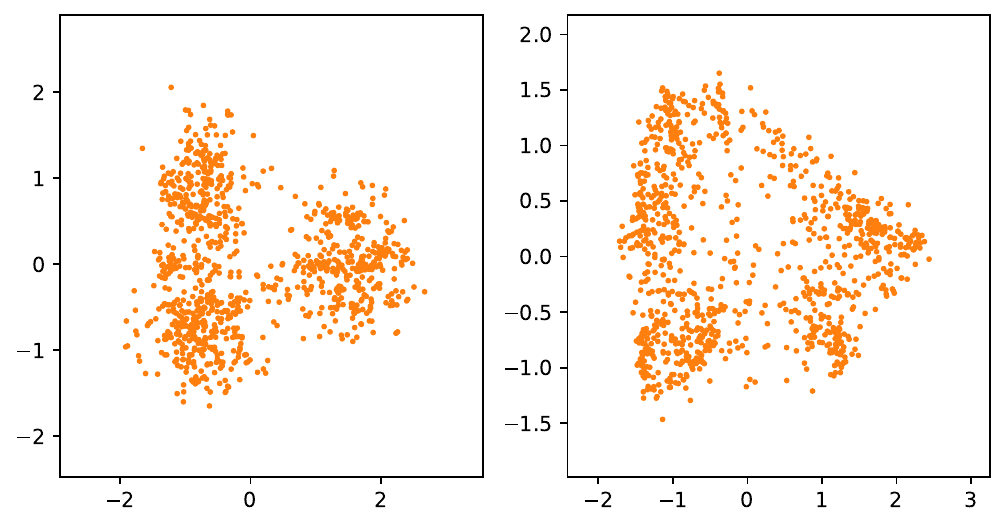}
        \caption{MM-B}
    \end{subfigure}
    \begin{subfigure}{0.40\columnwidth}
        \includegraphics[width=\textwidth]{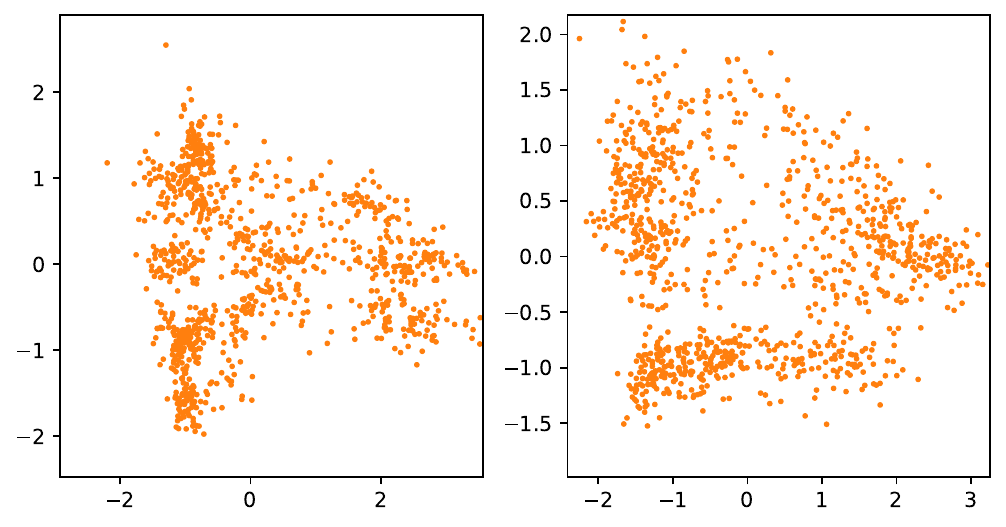}
        \caption{HOTET}
    \end{subfigure}
        \hfill
    \begin{subfigure}{0.40\columnwidth}
        \includegraphics[width=\textwidth]{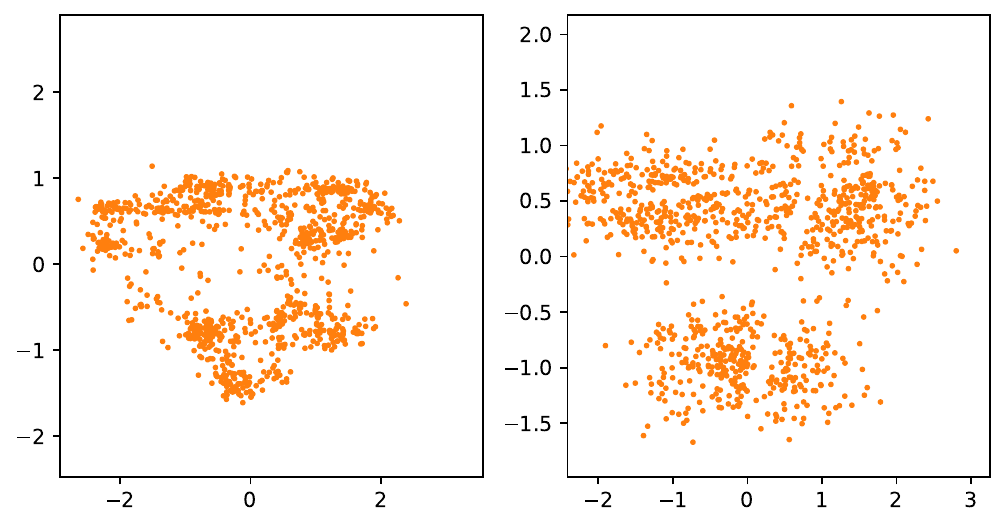}
        \caption{Meta OT}
    \end{subfigure}
    \caption{Visualization of the transported data mapped via various methods. The ground truth data are shown in the top-left corner. 
    For the other figures, the left (resp.\ right) subplots are generated by applying the forward maps \(\hat{T}\) (resp.\ inverse maps \(\hat{T}^{-1}\)) to the corresponding ground truth data. \label{fig:w2b_map}}
  \vspace{-1.5em}
\end{figure}

\subsection{W2B Benchmark}
\label{subsec:W2B}

We trained HOTET with the MM-B solver using the high-dimensional distribution dataset from W2B
For the estimated forward and inverse transport maps, we examine the divergence between \(\nu\) and \(\hat{T}_\pushforward\mu\), as well as the one between \(\mu\) and, \(\hat{T}^{-1}_\pushforward\nu\),
which are visualized in \Cref{fig:w2b_map}. 
To make the comparison fair, the architectures of the ICNNs used in all the methods are identical. 

The results are reported in \Cref{tab:uvp_w2b,tab:cos_w2b}, 
indicating that the transport maps generated by HOTET are comparable to those trained directly. 
This evidence validates that HOTET is able to generate quality transport maps.



\subsection{Predict OT Maps with HOTET}
\label{subsec:MGD}

We then evaluated the prediction performance of HOTET.
In this experiment, Gaussian mixtures with three components, across multiple dimensions are generated, and each with different mean and covariance. 
One mixture \(\nu\) was chosen as the reference, while 500 others formed the training set. 
After fitting 500 OT maps under HOTET, we then predicted the OT maps between \(\nu\) and 100 new Gaussian mixtures 
(we found that MMv2 performed better in this setting than MM-B). 
For comparison, Meta OT was also trained in the setting above.

The results are summarized in \Cref{tab:uvp_mix_gaussians}. 
The prediction performance of HOTET quickly exceeds baselines as the dimension increases,
exhibiting the capability of HOTET to capture distribution embeddings.
Due to space limit, evaluation on time efficiency is deferred to Appendix~\ref{app:time cost}.




\textbf{Key note: removal of the pretraining stage.} 
\cite{korotin2021neural} suggested a pre-training stage to turn the ICNN into an identity map, before the regular training procedure. This process is time-consuming and unnecessary given the recent advance in transfer learning;
we followed \cite{houlsby2019parameter} and initialized the weights $W_h \sim \mathcal{N}(0,0.1)$ in hypernetworks with a small variance. This way, we utilized the residual connection within ICNNs (with the tiny initial weights the ICNN already approaches an identity map) and thus can skip the pre-training stage. 

\begin{figure}[t]
  \centering
    \includegraphics[width=0.5\columnwidth]{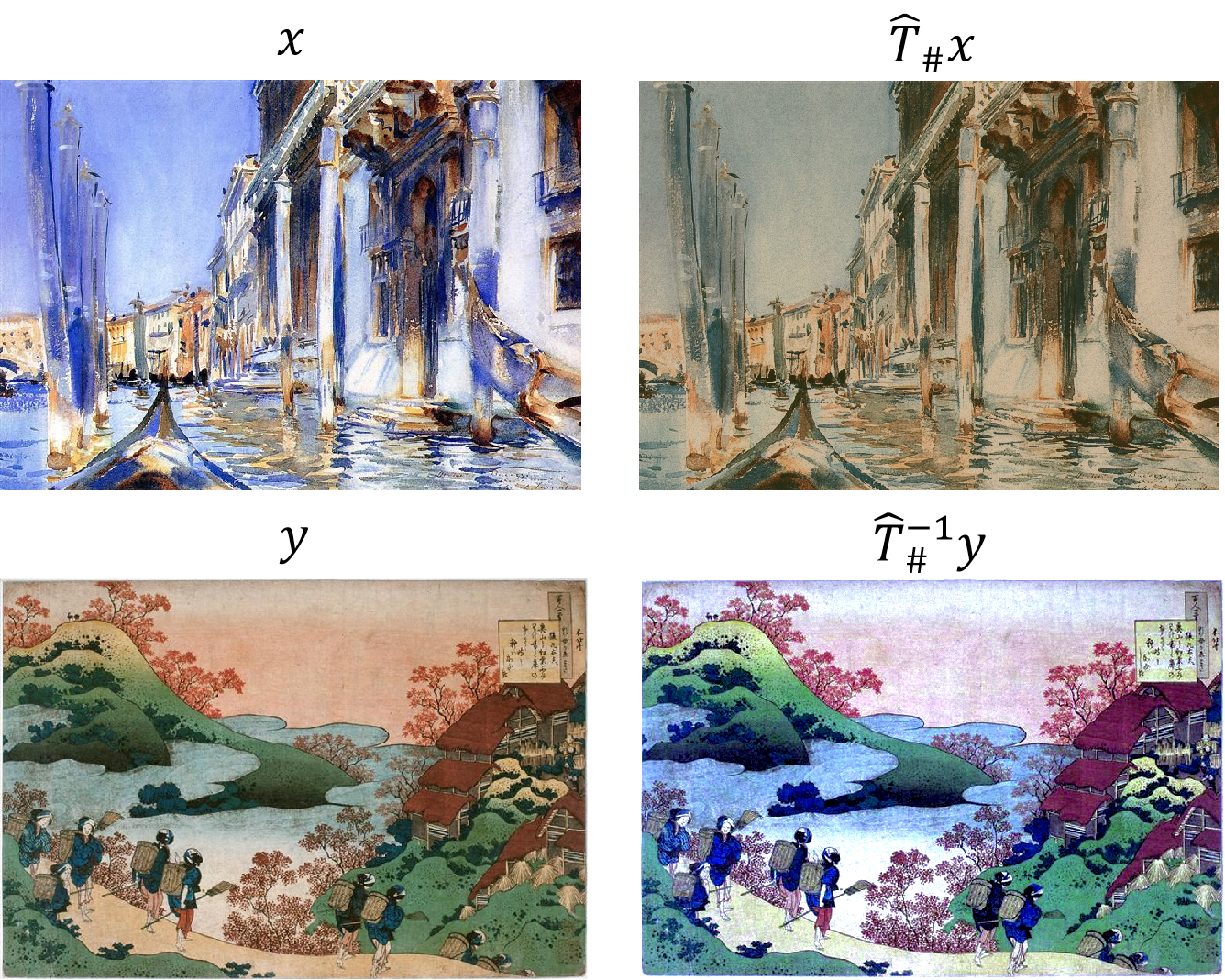}
        \caption{One-to-one color transfer using HOTET. \label{fig:one_color_transfer}}
  \vspace{-1em}
\end{figure}

\subsection{Color Transfer}
\label{subsec:color_transfer}
In this experiment, we used the OT map to transform the color histogram of an image to match that of another. The transport maps were constructed by interpreting the input RGB images as samples from their respective color distributions over the support \([0,1]^3\). 
As before, we considered both one-to-one and many-to-one settings, employing the MM-B solver.

\textbf{One-to-one color transfer.} Given two color histograms, $x$ and $y$, we trained HOTET to generate OT maps \(\hat{T},\hat{T}^{-1}\). To obtain new images, we replaced the colors of individual pixels so that the resulting color histograms became \(x_\text{trans}=\hat{T}_\pushforward x\) and \(y_\text{trans}=\hat{T}^{-1}_\pushforward y\), respectively. The results are shown in \Cref{fig:one_color_transfer}. 

\begin{figure}[t]
  \centering
    \includegraphics[width=0.65\columnwidth]{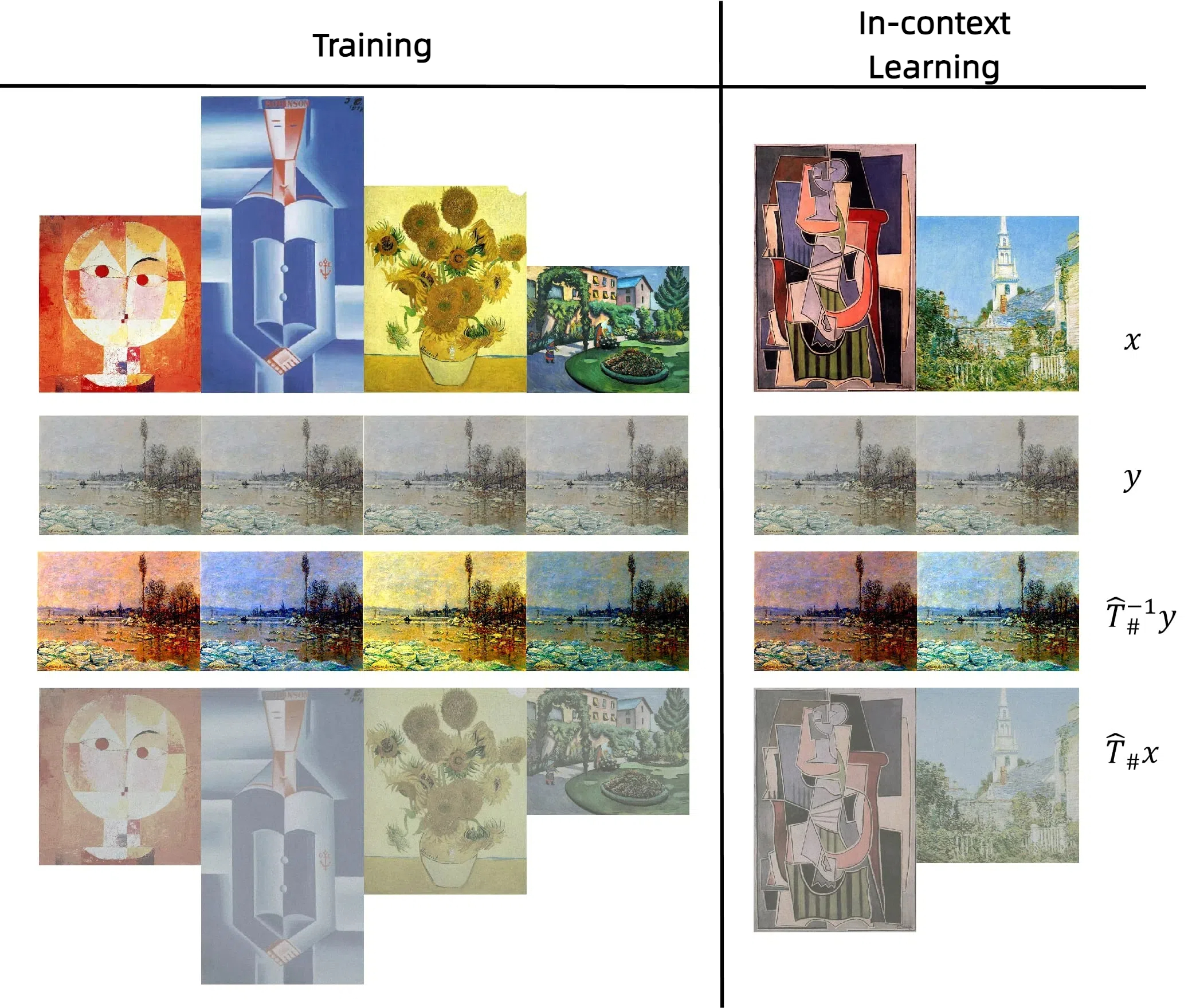}
    \caption{Multiple-to-one color transfer using HOTET. Showing both the training and in-context learning stages. \label{fig:multi_color_transfer}}
\end{figure}

\textbf{Multiple-to-one color transfer.} 
Given a collection of color histograms $X=\{x_1,x_2,\dotsc,x_n\}$ and a reference histogram $y$, we trained HOTET to generate simultaneously the forward transport maps \(\{\hat{T}_1,\hat{T}_2,\dotsc,\hat{T}_n\}\) and inverse maps \(\{\hat{T}_1^{-1}, \hat{T}_2^{-1}, \dotsc, \hat{T}_n^{-1}\}\). The images with new color histograms, shown in the left panel of \Cref{fig:multi_color_transfer}, were generated in the same way as in the one-to-one setting.

\textbf{In-context learning.} To further evaluate the generalization capacity of a trained HOTET, we followed the in-context learning setting in \cite{amos2022meta} and reused the trained HOTET model from the multiple-to-one setting to generate OT maps for previously unseen color histograms $x_\text{new}$. 
Instead of a full training run (5000 iterations), we fine-tuned the HOTET on new images through only 50 warm-up steps. 
The results are presented in the right panel of \Cref{fig:multi_color_transfer}, demonstrating that the OT maps produced by short-term fine-tuning are visually comparable to those generated through complete training.


\subsection{Ablation Studies on the Distribution Embedding Module}
\label{subsec:ablation} 


In this experiment, we \textbf{removed} the embedding module to investigate whether the overall performance would be impacted. 
We followed the setting in \Cref{subsec:MGD} and the dimension is set at 64;
the results are provided in \Cref{tab:ablation}. 
As expected, HOTET significantly outperforms the version without embedding, highlighting the importance of the embedding module in effectively capturing signals from empirical distributions.

\begin{table}[t]
    \centering
    \caption{$\calL^2$-UVP (\%) results of the two different methods for obtaining the forward and inverse transport maps. \label{tab:ablation}}
    \resizebox{0.6\columnwidth}{!}{
        \setlength{\tabcolsep}{3pt}
        \begin{tabular}{c|cc|cc}
            \toprule
            \multirow{2}{*}{Method} & \multicolumn{2}{c|}{Train} & \multicolumn{2}{c}{Predict} \\
            \cline{2-5}
              & \textit{Fowrad Map} & \textit{Inverse Map} & \textit{Fowrad Map} & \textit{Inverse Map} \\
\midrule
HOTET & 29.18 {\textcolor{gray}{\tiny ±0.46}} & 26.44 {\textcolor{gray}{\tiny ±0.36}} & 28.29 {\textcolor{gray}{\tiny ±0.83}} & 26.49 {\textcolor{gray}{\tiny ±0.35}} \\
No Emb. & 82.75 {\textcolor{gray}{\tiny ±0.55}} & 69.42 {\textcolor{gray}{\tiny ±0.47}} & 83.71 {\textcolor{gray}{\tiny ±0.51}} & 69.54 {\textcolor{gray}{\tiny ±0.64}} \\
\bottomrule
        \end{tabular}
    }
\vspace{-1em}
\end{table}

\section{Conclusions and Limitations}

In this paper, we recognize the increasing needs for computation of OT maps in modern signal processing, and propose a training paradigm HOTET to learn the OT maps between an unseen empirical distribution and a reference measure. 
In HOTET, information from the input distributions is extracted by a transformer, and then passed to a hypernetwork to generate the desired OT maps. 
Extensive experiments were conducted to demonstrate the efficacy of our new paradigm, showing that it is capable of producing quality OT maps.

\paragraph*{Limitations} 
Despite the efficacy of HOTET, it faces challenges in the many-to-many setting, where data are multiple i.i.d. distribution pairs~$(\mu_i, \nu_i)$. 
Further exploration is needed to address this complex scenario.

\section{Acknowledgments}
Mingchen Jiang was supported by 
the “R\&D Hub Aimed at Ensuring Transparency and Reliability of Generative AI Models” project of the Ministry of Education, Culture, Sports, Science and Technology.








\clearpage
\bibliographystyle{IEEEtran}
\bibliography{ref_full}








\appendices
\clearpage

\onecolumn

\bigskip
\begin{center}
{\LARGE\bf Appendix to ``\mytitle''}
\end{center}
\numberwithin{equation}{section}
\bigskip
\spacingset{1.5}

\section{More on Experiments}

\subsection{Implementation details and discussions}
\label{sec:imple-details}

Despite the theoretical results for OT, there are a few numerical issues in implementing the HOTET framework,
which significantly impact the empirical performance without proper practical adaptation.

\subsubsection{Asymmetry in computing multiple OT maps w.r.t. one reference measure}
In the setting of \Cref{sec:hotet-multi}, the roles of the two hypernetworks differ:
the hypernetwork $\calG$ for inverse maps still depends on $\mu_i$'s, and is expected to precisely transport the single reference measure $\nu$ onto different destinations.
Due to the inherent asymmetry in the MMv2 solver, practically the forward maps $\varphi_i$ usually show superior performance over the inverse ones.


We take the special characteristics into consideration when devising the implementation of HOTET. 
Specifically, for the OT maps required in downstream applications, such as Wasserstein embedding or color transfer, we suggest setting them as the forward maps, to obtain high-quality neural maps.





\subsubsection{Restriction on parameters of ICNN}

Recall from \Cref{sec:prelim_icnn} that the parameters of certain weight matrices in \Cref{eqn:icnn} must remain non-negative. 
To enforce this restriction, we adopt \emph{projected gradient descent} and formally apply and additional ReLU activation to the selected outputs of the hypernetwork; 
other alternatives, such as Softplus, are evaluated while we found ReLU is numerically the most stable choice.

\subsection{Implementation of the Base OT Solvers}
\label{app:base_solver}

In this work, we considered two base OT solvers: MM-B \cite{mallasto2019q} MMv2 \cite{pmlr-v119-makkuva20a}. Both solvers are proven to perform reasonably well via benchmarking \cite{korotin2021neural}. 

\paragraph{MM-B Solver} The MM-B solver is built upon a reformulation \cite[Thm.~2.9]{villani2003optimal} of the dual problem \eqref{eqn:DP}. By dropping the constant terms, the problem reduces to an equivalent formulation \cite[Eqn.~5]{korotin2021wasserstein}: 
\begin{equation}\label{eqn:wass_corr}
\min_{\varphi\in\mathcal{C}}\left\{\int\varphi(x)\,d\mu(x)+\int\varphi^\dagger(y)\,\dd\nu(y)\right\},\qquad\varphi^\dagger(y)\coloneqq\sup_{x\in\mb R^d}\left\{\inner{x,y}-\varphi(x)\right\}.\tag{Cor}
\end{equation}
where \(\varphi^\dagger\) is the Fenchel conjugate \cite{rockafellar1970} of \(\varphi\), and \(\mathcal{C}\) denotes the class of convex functions. The main challenge in computing \(\varphi^\dagger\) lies in finding \(x(y)\) that achieves the maximum for each \(y\) on the support of \(\nu\). The MM-B solver simplifies this process by addressing the inner problem only on minibatches sampled from \(\mu\) and \(\nu\). Specifically, let \(f_\theta\) be an ICNN, the MM-B solver, as implemented by \cite{korotin2021neural}, optimize \(\theta\) as follows: given minibatches \(X\sim\mu\) and \(Y\sim\nu\) of size \(B\), the loss function to be minimized is computed as \[L(\theta)=\frac{1}{B}\sum_{j=1}^Bf_\theta(X_j)-f_\theta\bigl(X_{i(j)}\bigr),\qquad i(j)\coloneqq\argmax_{i\in [B]}\inner{X_i, Y_j}-f_\theta(X_i).\] Although this approach produces a biased solution, it significantly accelerates computation.

To solve for \(\varphi\) and its conjugate simultaneously, the loss function can be symmetrized as \[L(\theta,\omega)=L(\theta)+\left(\frac{1}{B}\sum_{j=1}^Bg_\omega(Y_j)-g_\omega\bigl(Y_{k(j)}\bigr)\right),\qquad k(j)\coloneqq\argmax_{k\in [B]}\inner{Y_k, X_j}-g_\omega(Y_k).\] where \(g_\omega\) is an ICNN that approximates the conjugate of \(f_\theta\).

\paragraph{MMv2 Solver} The MMv2 solver, on the other hand, is based on the reformulation \cite[Thm.~3.3]{pmlr-v119-makkuva20a} \[\min_{\varphi\in\mathcal{C}}\int\varphi(x)\,d\mu(x)+\int\max_{\psi\in\mathcal{C}}\left\{\inner{\nabla\psi(y),y}-\varphi\circ\nabla\psi(y)\right\}\,\dd\nu(y).\] To compute the OT map, potentials \(\varphi\) and \(\psi\) are approximated by ICNNs \(f_\theta\) and \(g_\omega\), respectively. The parameters \(\theta\) and \(\omega\) are updated in an alternating fashion. For the inner problem, \(\omega\) is updated through maximizing \[L(\omega)=\frac{1}{B}\sum_{i=1}^B\inner{\nabla g_\omega(Y_i), Y_i}-f_\theta\circ\nabla g_\omega(Y_i)\] via SGD, with multiple iterations performed to ensure convergence. Afterward, \(\theta\) is updated by minimizing \[L(\theta)=\frac{1}{B}\sum_{i=1}^Bf_\theta(X_i)-f_\theta\circ\nabla g_\omega(Y_i).\] This procedure is repeated until both \(\theta\) and \(\omega\) converge to near-optimal values.


\subsection{Pseudocode for Network Training}
\label{app:training_algo}

\Cref{alg:training} outlines the training process for HOTET using the MM-B solver. Here, $\mathcal{X}\coloneqq\{X_i\}_{i=0}^{n}$ represents the set of distributions, and $Y$ is the reference distribution. In the case where $\mathcal{X}$ contains only a single distribution, it reduces to a classical one-to-one OT problem. For scenarios where \(n>1\) (e.g. OT maps prediction), a batch of distributions, with size $B\leq n$, is sampled from $\mathcal{X}$. Then, from each sampled distribution and the reference $y$, batches of size $b$ are drawn for evaluating the individual OT losses. The networks $f_\theta$ and $g_\omega$ are ICNNs that parameterize the convex dual potentials. The hypernetworks $\calF$ and $\calG$ generate the parameters for these potential networks, while $\calE$ serving as the embedding module. The variable $K$ denotes the total number of training iterations.


\begin{algorithm}[ht]
\caption{Training Procedure of HOTET with MM-B Solver \label{alg:training}}  
\begin{algorithmic}[1]  
\Procedure{TrainModel}{$\mathcal{X}, Y, f_\theta, g_\omega, \mathcal{F}, \mathcal{G}, \mathcal{E}$}  
    \State Initialize the parameters of each module  
    \For{\texttt{t} $= 1, \dotsc, K$}  
        \State Sample batch $\mathcal{X}_\text{batch}$ of size \texttt{B} from $\mathcal{X}$.  
        \State $\Loss \gets 0$  
        \For{$\texttt{i}=1, \dotsc, \texttt{B}$}  
            \State Sample batches \texttt{x}, \texttt{y} of size \texttt{b} from $X_i, Y$, respectively  
            \State \texttt{embedding\_x} $\gets \mathcal{E}(\texttt{x})$  
            \State \texttt{embedding\_y} $\gets \mathcal{E}(\texttt{y})$  
            \State $\Loss_{xy} \gets \Call{ComputeLossForward}{\texttt{x}, \texttt{y}, \mathcal{F}, \texttt{embedding\_x}}$  
            \State $\Loss_{yx} \gets \Call{ComputeLossInverse}{\texttt{y}, \texttt{x}, \mathcal{G}, \texttt{embedding\_y}}$  
            \State $\Loss \gets \Loss + (\Loss_{xy} + \Loss_{yx}) / (2 \cdot \texttt{B})$  
        \EndFor  
        \State Update model $\mathcal{E}, \mathcal{F}, \mathcal{G}$ according to $\Loss$  
    \EndFor  
\EndProcedure  

\State\phantom{A}  
\Function{ComputeLossForward}{$\texttt{x}, \texttt{y}, \mathcal{F}, \texttt{embedding\_x}$}   
    \State $\theta \gets \mathcal{F}(\texttt{embedding\_x})$  
    \State $\texttt{x\_push} \gets \nabla f(\texttt{x} \mid \theta)$  
    \State $\texttt{xy} \gets \inner{\texttt{x}, \texttt{y}}$  
    \State $\texttt{idx\_y} \gets \text{argmax}(\texttt{xy} - \texttt{x\_push}, \text{dim}=0)$  
    \State $\texttt{y\_push} \gets \texttt{x[idx\_y]}$  
    \State $\texttt{W\_loss\_xy} \gets \text{mean}(\texttt{x\_push} - \nabla f(\texttt{y\_push} \mid \theta))$  
\EndFunction  

\State\phantom{A}  
\Function{ComputeLossInverse}{$\texttt{y}, \texttt{x}, \mathcal{G}, \texttt{embedding\_y}$}   
    \State $\omega \gets \mathcal{G}(\texttt{embedding\_y})$  
    \State $\texttt{y\_push} \gets \nabla g(\texttt{y} \mid \omega)$  
    \State $\texttt{yx} \gets \inner{\texttt{y}, \texttt{x}}$  
    \State $\texttt{idx\_x} \gets \text{argmax}(\texttt{yx} - \texttt{y\_push}, \text{dim}=0)$  
    \State $\texttt{x\_push} \gets \texttt{y[idx\_x]}$  
    \State $\texttt{W\_loss\_yx} \gets \text{mean}(\texttt{y\_push} - \nabla g(\texttt{x\_push} \mid \omega))$  
\EndFunction  
\end{algorithmic}  
\end{algorithm}

\begin{table}[h]
    \caption{Time cost (sec) of training different models. The batch sizes in the 32/64 dimensional settings are 256, and in other settings are 1024. The GPU is a single Nvidia RTX 4090. \label{tab:time_efficiency}
    }
    \centering
        \begin{tabular}{c|cccc|cc}
            \toprule
            Model & 2 & 4 & 8 & 16 & 32 & 64 \\
            \midrule
            MetaOT & 1161 & 1183 & 1214 & 1248 & 1405 & 4198 \\
            HOTET-MMB & 1228 & 1239 & 1339 & 1405 & 1223 & 2324  \\
            HOTET-MMv2 & 30972 & 31762 & 29106 & 30541 & 28150 & 20120 \\
            MM-B & 20695 & 21848 & 24099 & 24615 & 19618 & 52633 \\
            MMv2 & 409284 & 365529 & 418475 & 411523 & 401511 & 395583 \\
            \bottomrule
        \end{tabular}
\end{table}

\subsection{Runtime analysis}
\label{app:time cost}

We further assessed the time efficiency of HOTET, MetaOT, and repeating MMv2 solver in an 8-dimensional setting. The results in \Cref{tab:time_efficiency} showed that HOTET performs similarly to MetaOT. However, training networks with MMv2 solver directly is more time-consuming, as it requires training each distribution pair individually 500 times in this setting. Meanwhile, a direct MMv2 solver does not have prediction capabilities, as the potentials of the trained networks only represents the transport maps between the two input distributions.

\subsection{Additional Experiment Results}
\label{subsec:w2b_appendix}

\begin{figure}[!htb]
    \centering
    \begin{subfigure}{0.48\textwidth}
        \includegraphics[width=\textwidth]{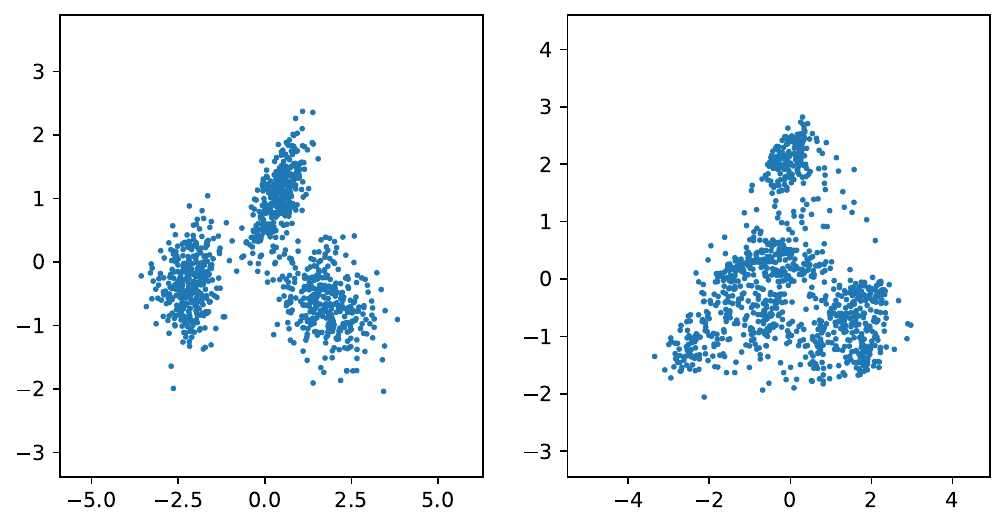}
        \caption{Ground Truth}
    \end{subfigure}
    \hfill
    \begin{subfigure}{0.48\textwidth}
        \includegraphics[width=\textwidth]{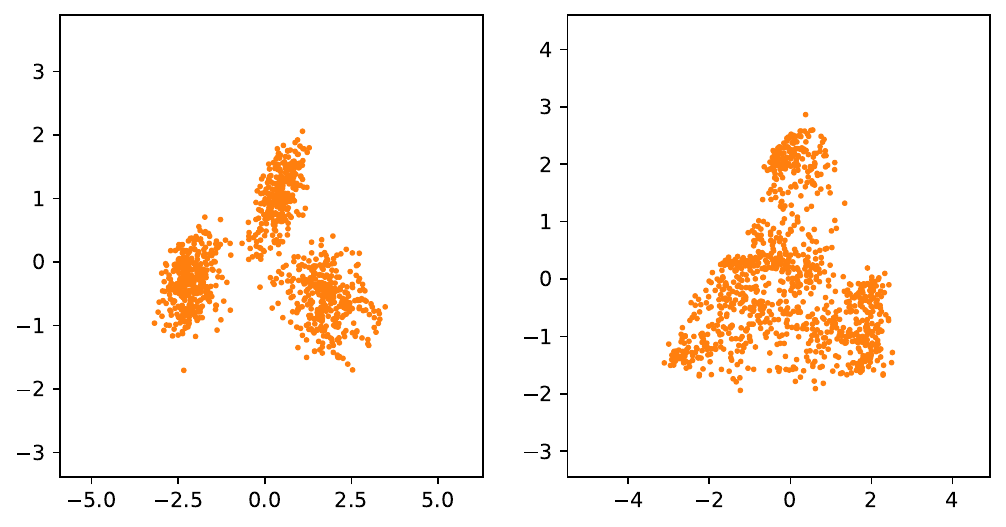}
        \caption{MM-B}
    \end{subfigure}
    \begin{subfigure}{0.48\textwidth}
        \includegraphics[width=\textwidth]{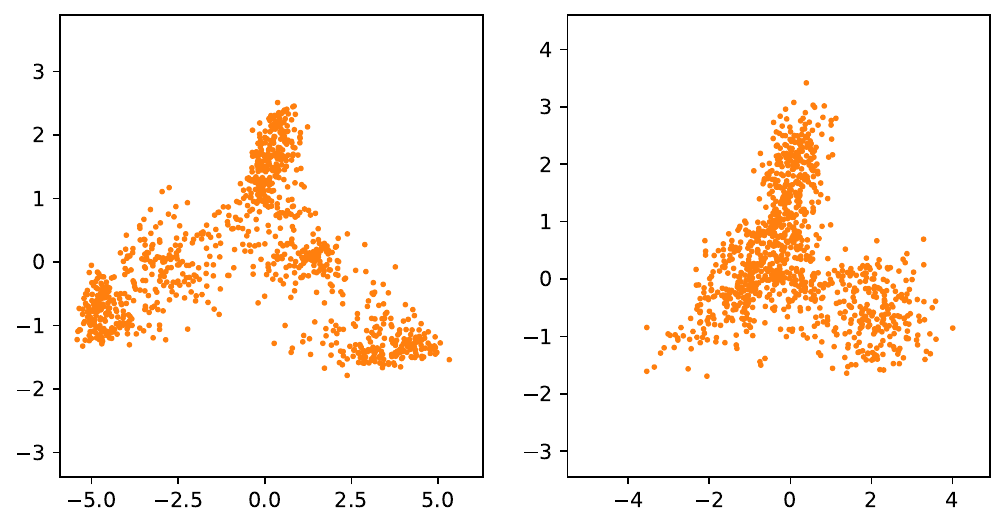}
        \caption{HOTET}
    \end{subfigure}
        \hfill
    \begin{subfigure}{0.48\textwidth}
        \includegraphics[width=\textwidth]{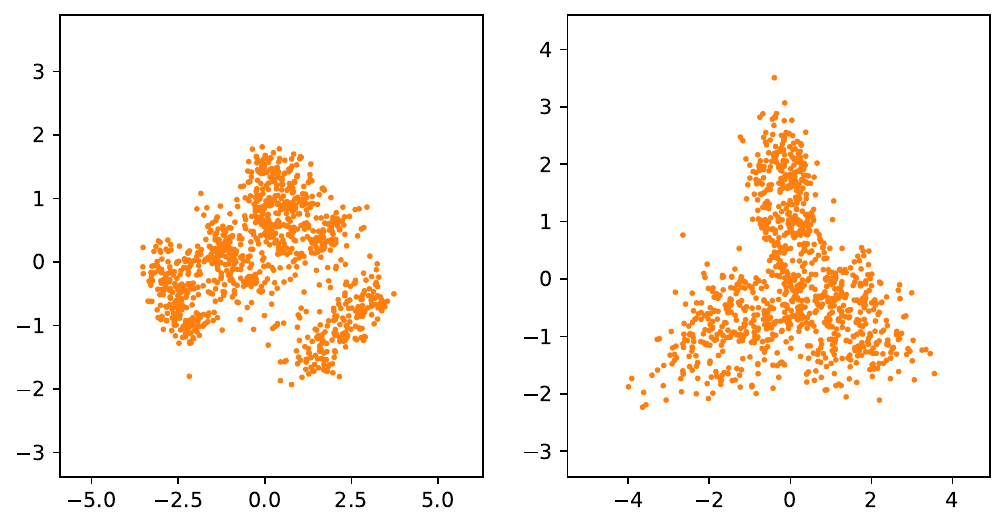}
        \caption{Meta OT}
    \end{subfigure}
    \caption{Samples generated by the forward and inverse maps in $d=4$, compared with ground truth input distributions.}\label{fig:w2b_dim4}
\end{figure}

\begin{figure}[!htb]
    \centering
    \begin{subfigure}{0.48\textwidth}
        \includegraphics[width=\textwidth]{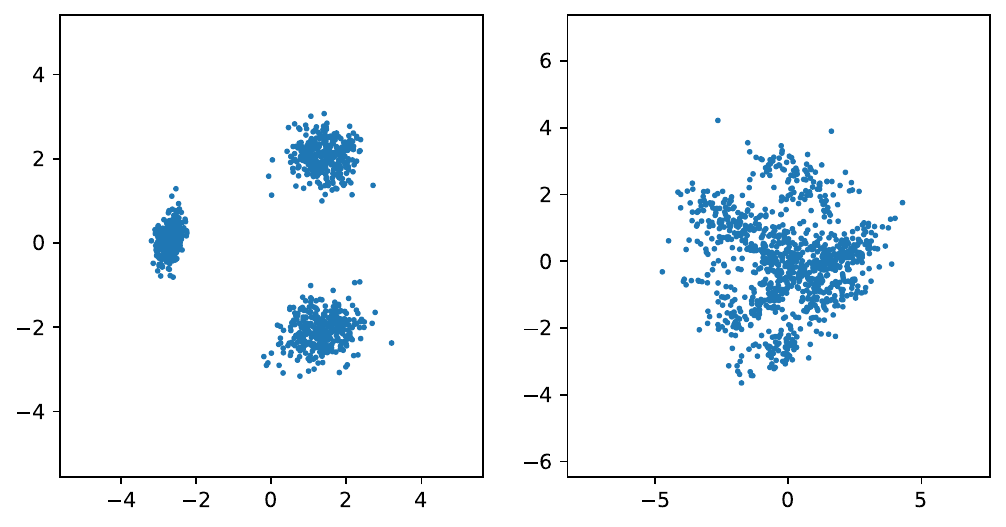}
        \caption{Ground Truth}
    \end{subfigure}
    \hfill
    \begin{subfigure}{0.48\textwidth}
        \includegraphics[width=\textwidth]{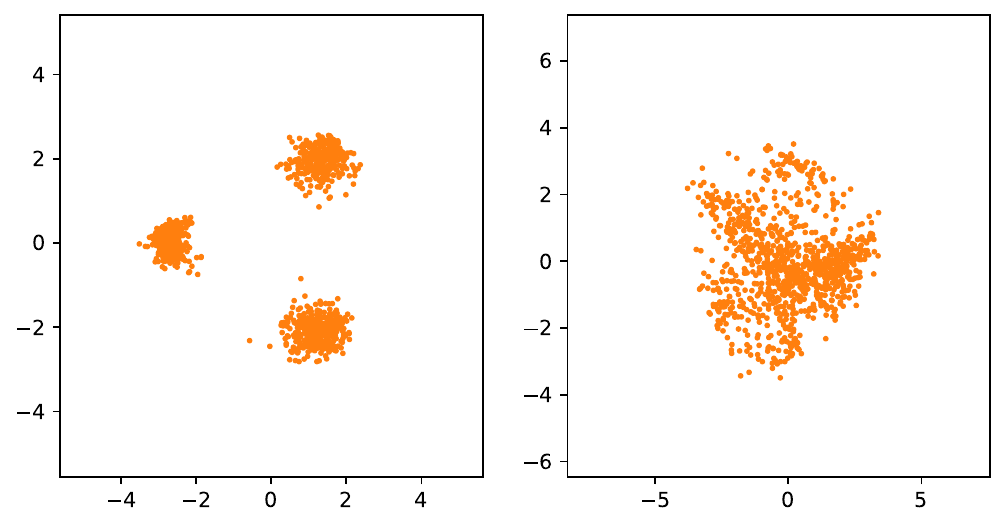}
        \caption{MM-B}
    \end{subfigure}
    \begin{subfigure}{0.48\textwidth}
        \includegraphics[width=\textwidth]{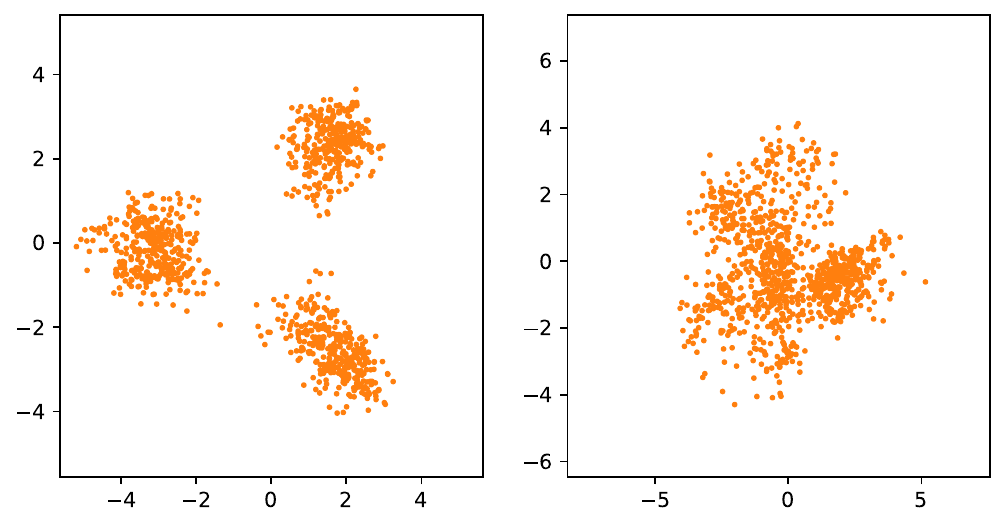}
        \caption{HOTET}
    \end{subfigure}
        \hfill
    \begin{subfigure}{0.48\textwidth}
        \includegraphics[width=\textwidth]{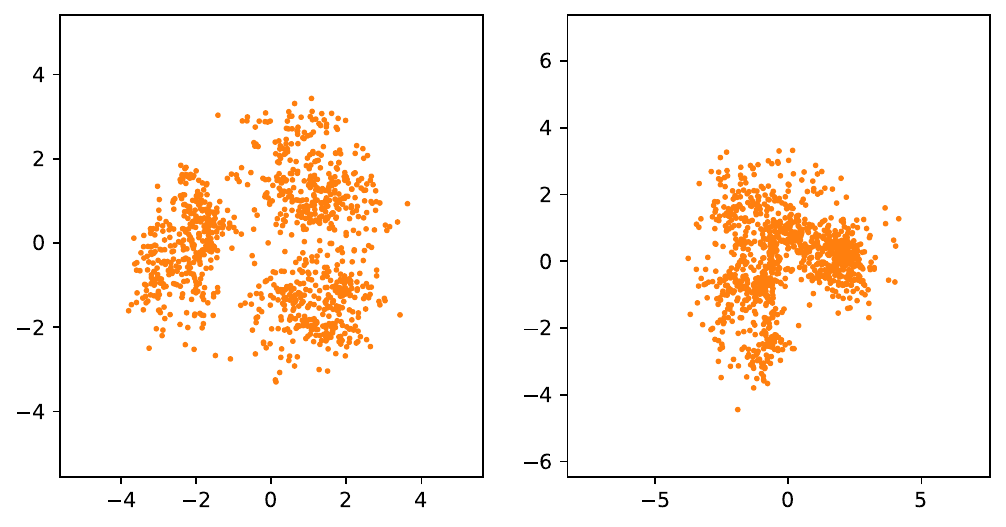}
        \caption{Meta OT}
    \end{subfigure}
    \caption{Samples generated by the forward and inverse maps in $d=8$, compared with ground truth input distributions.}\label{fig:w2b_dim8}
\end{figure}

\Cref{fig:w2b_dim4,fig:w2b_dim8} present OT maps learned in the W2B experiment (\Cref{subsec:W2B}) with $d=4$ and $d=8$. The results are projected onto the first 2 principal directions for visualization.

\subsection{Choosing the Solvers}
\label{app:solvers}
MM-B and MMv2 perform differently in the experiment settings in \Cref{subsec:W2B} and \Cref{subsec:MGD}, the details are in \Cref{tab:dif_w2b} and \Cref{tab:dif_multi}. Therefore, we choose the better performed one in our main paper.

\begin{table}[!t]
    \centering
    \caption{\label{tab:dif_w2b}
    Performance of the constructed forward (fwd) and inverse (inv) transport maps by MM-B and MMv2 solvers in W2B benchmark ($\calL^2$-UVP (\%) as the metric). 
    Lower implies the fitted map approximates the true OT map better. 
    }
    \begin{tabular}{c|cc|cc}
        \toprule
        \thead{DIM}  & \thead{HOTET-MMB (fwd)} & \thead{HOTET-MMv2 (fwd)} &\thead{HOTET-MMB (inv)} & \thead{HOTET-MMv2 (inv)} \\
\midrule
2 & 5.03 ± 0.33 & 13.34 ± 0.46 & 10.49 ± 0.52 & 5.31 ± 0.58 \\
4 & 10.06 ± 0.39 & 30.74 ± 0.55 & 15.54 ± 0.46 & 23.84 ± 0.67 \\
8 & 11.38 ± 0.41 & 33.90 ± 0.61 & 18.79 ± 0.45 & 16.92 ± 0.52 \\
16 & 16.91 ± 0.58 & 30.86 ± 0.56 & 25.62 ± 0.49 & 28.72 ± 0.39 \\
32 & 20.87 ± 0.56 & 40.72 ± 0.54 & 23.04 ± 0.53 & 35.71 ± 0.47 \\
64 & 37.62 ± 0.84 & 38.42 ± 0.51 & 22.94 ± 0.87 & 36.06 ± 0.62 \\
\bottomrule
    \end{tabular}
\end{table}

\begin{table}[!t]
    \centering
    \caption{\label{tab:dif_multi}
    Performance of the constructed forward (fwd) and inverse (inv) transport maps by MM-B and MMv2 solvers in predicting OT maps setting ($\calL^2$-UVP (\%) as the metric). 
    Lower implies the fitted map approximates the true OT map better. 
    }
    \resizebox{0.90\textwidth}{!}{
    \begin{tabular}{c|cccc|cccc}
        \toprule
        \multirow{2}{*}{\thead{DIM}} & \multicolumn{4}{c|}{\thead{Train}} & \multicolumn{4}{c}{\thead{Predict}} \\
        \cline{2-9}
          & \thead{HOTET-MMv2 (fwd)} & \thead{HOTET-MMB (fwd)} & \thead{HOTET-MMv2 (inv)} & \thead{HOTET-MMB (inv)} & \thead{HOTET-MMv2 (fwd)} & \thead{HOTET-MMB (fwd)} & \thead{HOTET-MMv2 (inv)} & \thead{HOTET-MMB (inv)} \\
\midrule  
2 & 3.25 ± 0.56 & 4.23 ± 0.54 & 3.01 ± 0.49 & 4.03 ± 0.52 & 3.13 ± 0.47 & 4.12 ± 0.62 & 3.07 ± 0.45 & 3.84 ± 0.61 \\
4 & 3.40 ± 0.29 & 4.95 ± 0.61 & 3.44 ± 0.30 & 3.73 ± 0.29 & 3.37 ± 0.27 & 3.57 ± 0.33 & 3.43 ± 0.29 & 3.66 ± 0.36 \\
8 & 6.59 ± 0.28 & 6.95 ± 0.28 & 6.48 ± 0.27 & 7.00 ± 0.28 & 6.54 ± 0.31 & 6.80 ± 0.20 & 6.45 ± 0.28 & 6.94 ± 0.22 \\
16 & 10.72 ± 0.26 & 12.60 ± 0.26 & 11.19 ± 0.27 & 12.64 ± 0.31 & 10.59 ± 0.25 & 12.66 ± 0.28 & 11.05 ± 0.25 & 12.71 ± 0.31 \\
32 & 18.00 ± 0.47 & 20.36 ± 0.26 & 18.96 ± 0.54 & 20.35 ± 0.25 & 18.03 ± 0.44 & 20.00 ± 0.22 & 19.00 ± 0.53 & 19.93 ± 0.21 \\
64 & 29.18 ± 0.46 & 28.69 ± 0.18 & 26.44 ± 0.36 & 28.72 ± 0.18 & 28.29 ± 0.83 & 28.74 ± 0.18 & 26.49 ± 0.35 & 28.84 ± 0.17 \\
\bottomrule
    \end{tabular}
    }
\end{table}

\subsection{Validating the Embedding Module with the MNIST Dataset}
\label{subsec:MNIST}
\begin{figure}[H]
  \centering
  \includegraphics[width=0.7\columnwidth]{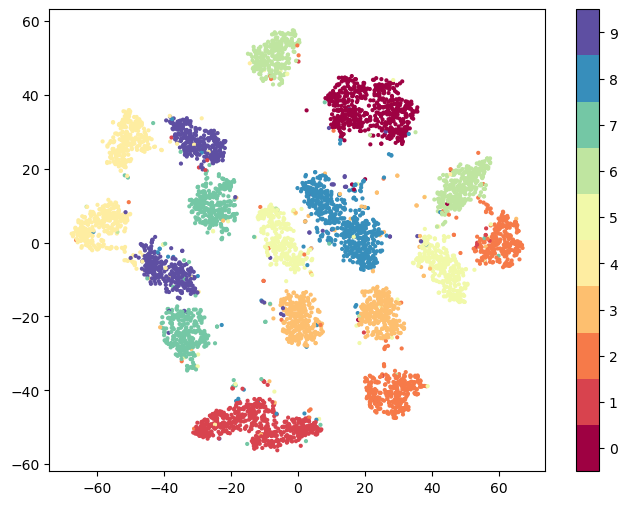}
  \caption{A $t$-SNE visualization of the 128-dimensional vectors produced by the embedding module for MNIST, with the true labels used to color the points. The embeddings of the same digit clearly clustered together, indicating that the embedding module effectively preserved the information from the original images. \label{fig:embd_tsne}}
\end{figure}

In this experiment, we examined the embeddings produced by the embedding module \(\calE\) from a HOTET trained on the MNIST~\cite{lecun1998gradient} handwritten digits dataset. We selected 6000 images from the training set (600 per digit), and then trained a VAE~\cite{kingma2013auto} to map them to 6000 \(3\)-dimensional latent distributions. Next, we trained a HOTET with embedding dimension \(d=128\) to learn OT maps between the latent distributions and the reference measures \(N(\bm{0}, \bm{I}_3)\). Afterwards, we visualized the training set embeddings using \(t\)-SNE to assess whether the embedding module effectively captured the information from the original images. The result is showed in \Cref{fig:embd_tsne}.




\subsection{Figure Result of Individual OT Maps}

\section{Useful Facts}
\label{app:useful_facts}
\providecommand{\upGamma}{\Gamma}
\providecommand{\uppi}{\pi}

\subsection{ICNN}
\label{app:icnn}

The fully connected ICNN is a feed-forward neural network whose intermittent layer \( z_{\ell} \) is the activation of the linear transformation of the previous layer, plus the affine transformation of the input \(z_0\coloneqq x\). 
In other words, for \(\ell\in\{1,\dotsc,L\}\),
\begin{align}
z_{\ell}\coloneqq \sigma_\ell(A_{\ell-1} z_{\ell-1} + W_{\ell-1} x + b_{\ell-1}),
\label{eqn:icnn}
\end{align}
where \(A\) is a non-negative matrix, $W, b$ are regular unrestricted weight matrix and bias vector, and \(\sigma\) is a convex and non-decreasing activation function. 
The final ICNN output is \(\Tilde{\varphi}(x) = z_L \) for some pre-specified \( L \geq 1 \).

The structure of ICNN is justified by the following facts:
\begin{itemize}[leftmargin=*]
\item The composition of a convex and non-decreasing function and a convex function is convex.
\item The composition of a convex function and an affine function is convex.
\item The non-negative sum of convex functions is also convex.
\end{itemize}


\subsection{Transformer}
\label{app:transformer_icnn}

Transformers \cite{NIPS2017_3f5ee243} are a type of neural network architecture primarily used in natural language processing (NLP) tasks. They are composed by \(L\) stacked layers, where each layer comprises of a multi-headed attention and a fully connected feed-forward network (FFN) sub-layer. The attention sub-layer, assuming \(h\) heads and dimension size \(d\) for each head, first maps an input \(X \in \mathbb{R}^{n \times hd}\) into the query (\(Q\)), key (\(K\)), and value (\(V\)) matrices through the following affine transformations:
\begin{equation}
\label{eqn:attn-weights}
Q/K/V = XW_{[q/k/v]} + \bm{1}b^\top_{[q/k/v]},
\end{equation}
where \(Q, K, V \in \mathbb{R}^{n \times hp}\), \(W_q, W_k, W_v\) are \(hd \times hd\) weight matrices, 
and \(b_q, b_k, b_v \in \mathbb{R}^{N_hp}\) are the bias terms \footnote{To ease the notations we adopt the setting where \(X, Q, K, V\) have the same shape.}. After the transformation, the three 
components \(Q, K, V\) are split into \(h\) blocks corresponding to different heads. For example, 
\(Q\) is re-written as \(Q = (Q^{(1)}, \ldots, Q^{(N_h)})\), where each block 
\(Q^{(h)} = XW_q^{(h)} + 1(b_q^{(h)})^T\) is an \(n\times p\) matrix, and \(W_q^{(h)}, b_q^{(h)}\) are the corresponding 
parts in \(W_q, b_q\). The attention output for the \(h^{th}\) head is then computed as:
\begin{equation}\label{eqn:attn}
\mtx{L}^{(h)} \mtx{V}^{(h)} \defeq \operatorname{softmax}(\mtx{Q}^{(h)} (\mtx{K}^{(h)})^T / \sqrt{p}) \mtx{V}^{(h)}= (\mtx{D}^{(h)})^{-1} \mtx{M}^{(h)} \mtx{V}^{(h)},
\end{equation}
where \(M^{(h)}\coloneqq\exp(Q^{(h)}(K^{(h)})^T / \sqrt{p})\) and \(D^{(h)}\) is a diagonal matrix in which \(D_{ii}^{(h)}\) is the sum of the \(i\)-th row in \(M^{(h)}\), corresponding to the normalization part in softmax.

After we obtain the outputs in each head, they are concatenated as,
\begin{equation}
L\coloneqq\left(L^{(1)}V^{(1)}, \ldots, L^{(N_h)}V^{(N_h)}\right),
\end{equation}
followed by the overall output,
\begin{equation}
LW_o + 1b_o^T,
\end{equation}
where \(W_o\) and \(b_o\) are similarly sized as the other matrices in \Cref{eqn:attn-weights}.

\subsection{Attention as Kernel Estimators}
\label{sec:attn-as-kernel}

For each head in the attention module, we have given the expression of attention output in Equation~\ref{eqn:attn}.
In this subsection, we will re-write attention as a kernel estimator to show the connection.

In computing the attention output (of a single head), we have an input sequence $\{x_i\}_{i=1}^n$ (the rows in $X$) and accordingly we can obtain $N$ \footnote{Note that $N$ may not always equal $n$, such as in cross attention ($N \neq n$) or in prefix-tuning ($N > n$ due to the prefix pretended to the key matrix)~\cite{Li2021PrefixTuningOC}.} key vectors $\{k_j\}_{j=1}^N \subset \mb R^p$ (from the \textit{key} matrix $\mtx{K}$) and query vectors $\{q_i\}_{i=1}^n \subset \mb R^p$ (from $\mtx{Q}$).\footnote{In this subsection we omit the superscript $(h)$ for simplicity since the discussion is limited within a single head}
The original goal of self-attention is to obtain the representation of each input token $x_i$: $g(x_i)$.
By denotation exchange: $q_i\coloneqq x_i$ and $f(q_i)\coloneqq g(x_i)$, we can also understand the aforementioned self-attention module as returning the representation $f(q_i)$ of the input query vector $q_i$ through $\{k_j\}_{j=1}^n$,
which behaves as a kernel estimator~\cite{choromanski2020rethinking, chen2021skyformer}.
Specifically, for a single query vector $q_i$, a Nadaraya--Watson kernel estimator~\cite[Definition~5.39]{wasserman2006all} models its representation as,
\begin{align}
\label{eq:kernel_estimator}
f(q_i) = \sum_{j=1}^n \ell_j(q_i) c_j,~\quad\text{where}\quad \ell_j(q_i) \coloneqq \frac{\kappa(q_i, k_j)}{\sum_{j'=1}^N \kappa(q_i, k_{j'})}.
\end{align}

Here, $\kappa(\cdot, \cdot)$ is a kernel function, and $c_j$'s are the coefficients ($c_j$ can either be a scalar or a vector in different applications) that are learned during training. In this estimator, $\{k_j\}_{j=1}^n$ serve as the \textit{supporting points} which help construct the representation for an input $q_i$.

For kernel function $\kappa(x, y) = \exp\left(\dotp{x}{y} / \sqrt{p} \right)$,
we slightly abuse the notation $\kappa(\mtx{Q}, \mtx{K})$ to represent an $n$-by-$N$ empirical kernel matrix, whose element in the $i$-th row and the $j$-th column is $\kappa(q_i, k_j), \forall i \in [n], j \in [N]$.
With these notations, the representation of the transformed output will be,
\begin{align}
\label{eq:matrix_form}
\mtx{D}^{-1} \kappa(\mtx{Q}, \mtx{K}) \mtx{C},
\end{align}
where $\mtx{D}$ is a diagonal matrix for row normalization in Eq.~(\ref{eq:kernel_estimator}), and $\mtx{C}$ is an $N$-by-$p$ matrix whose $j$-th row is $c_j$.


\subsection{Potential Pathways to the Incorporation of Sample Weights into Distribution Embedding}
\label{app:sample-weights}

Considering the correspondence between Equation~(\ref{eq:matrix_form}) and the standard softmax attention in Equation~(\ref{eqn:attn}),
we are indeed able incorporate the sample weights in an empirical distribution to attention (as mentioned in \Cref{sec:layer_embedding}). 
The new character will allow a transformer to embed arbitrary empirical distributions and generate layer embeddings for hypernetworks.


Originally in transformers, all the tokens are assumed to share the equal weights, while a general empirical distribution allows non-uniform atom masses.
To address the issue, we can make an analogy between self-attention and the Nadaraya--Watson kernel estimator in \Cref{eq:kernel_estimator}. 

We first rewrite $\ell_j(q_i)$'s in \Cref{eq:kernel_estimator} to highlight the implicitly assumed uniform sample weights:
\begin{align*}
\ell_j(q_i) = \frac{\kappa(q_i, k_j)}{\sum_{j'=1}^N \kappa(q_i, k_{j'})}
= \frac{\frac1N \cdot \kappa(q_i, k_j)}{\sum_{j'=1}^N \frac1N \cdot \kappa(q_i, k_{j'})},
\end{align*}
which allows an immediate extension in the weighted case;
given the normalized sample weights $\mtx m = \{m_1, m_2, \dotsc, m_n\}$ with $\sum_{j=1}^N m_j = 1$, we can modify the coefficients $\ell_j(q_i)$'s in \Cref{eq:kernel_estimator} as
\begin{align*}
\ell_j(q_i) = \frac{m_j \cdot \kappa(q_i, k_j)}{\sum_{j'=1}^N m_{j'} \cdot \kappa(q_i, k_{j'})}.
\end{align*}
Ultimately, the weighted attention is expressed as follows:
\begin{align}
\mtx D^{-1} \exp\paren{\frac{\mtx Q \mtx K^T}{\sqrt{p}}} \diag\paren{N \mtx m} \mtx V \nonumber = \softmax\paren{\frac{\mtx Q \mtx K^T}{\sqrt{p}} \diag\paren{\ln{N \mtx m}}} \mtx V,
\end{align}
where the row normalization matrix $\mtx D$ is reloaded as $\diag\paren{\exp\paren{\frac{\mtx Q \mtx K^T}{\sqrt{p}}} N \mtx m}$.

For the output $\mtx H$ of the whole transformer, we can apply a weighted average pooling to obtain the final embedding:
\begin{align}
\mtx z = \mtx H^T \mtx m.
\end{align}
The embedding $\mtx z$ will then be passed to the hypernetwork for generating the transport map.

\subsection{Architecture Comparison with Existing Methods} 
\label{app:comparison}

We noted that our proposed paradigm shared some similarities with the general ideas of \textsc{CondOT} and Meta OT. Therefore, we shall briefly discuss the differences between our methods and theirs.

The setting of \textsc{CondOT} is based on a regression formulation. Therefore, the input source and target measures must be paired. Our scheme is more flexible compared to theirs in this regard, as it allows the numbers of source and target measures to differ. Further, the transport maps in \textsc{CondOT} are modulated by having an additional context variable as an input, while our transport maps are generated from properly trained hypernetworks. Lastly, we devised an end-to-end pipeline to extract information from source measures by transformers, while \textsc{CondOT} relied on externally given information to obtain the context variables. 

The Meta OT model, on the other hand, makes use of hypernetworks to generate the transport maps. Therefore, it is more comparable to our proposed method. In Meta OT, however, the distributions are passed directly to the hypernetworks as inputs, which means they must be concatenated, padded, or resized if their size mismatch. The transformer module in HOTET resolves this matter and extract the information more efficiently, as explained in \Cref{sec:layer_embedding}.



\section{Miscellanies}



\subsection{Notations}
\label{sec:notation}

We denote by \(\calE\) the embedding module, and \(\calF,\calG\) the hynernetworks in HOTET, respectively. Given distributions \(\mu\) and \(\nu\), we use \(T_{\mu\to\nu}\) to denote the true OT map that pushforwards \(\mu\) to \(\nu\), and omit the subscript when the context is clear. Accordingly, we use \(\varphi,\psi\) to denote the potential functions and \(f, g\) to denote the networks used for approximating \(\varphi,\psi\).

\subsection{Hyperparameters}
\label{sec:hyperparam}
For the detail settings, \Cref{tab:hyper_param_w2b,tab:hyper_param_mix,tab:hyper_param_color} show the hyperparameters we used in our experiments.


\begin{table}[H]
    \centering
    \begin{tabular}{c|ccc}
    \toprule
    Model & LR & Batch Size & Total Iterations \\
    \midrule
    HOTET & $10^{-3}$ & 1024 & 5000 \\
    MetaOT & $10^{-3}$ & 1024 & 5000  \\
    MM-B & $10^{-3}$ & 1024 & 5000\\
    \bottomrule
    \end{tabular}
    \caption{Hyperparamter in the W2B experiment.\label{tab:hyper_param_w2b}}
\end{table}

\begin{table}[H]
    \centering
    \begin{tabular}{c|cccc}
    \toprule
    Model & LR & Batch Size (data) & Batch Size (distributions) & Total Iterations \\
    \midrule
    HOTET & $10^{-3}$ & 1024 (dim=2,4,8,16) / 256 (dim=32, 64)& 8 & 5000 \\
    MetaOT & $10^{-3}$ & 1024 (dim=2,4,8,16) / 256 (dim=32, 64)& 8 & 5000  \\
    MMv2 & $10^{-3}$ & 1024 & N/A & 5000\\
    \bottomrule
    \end{tabular}
    \caption{Hyperparamter in the OT maps prediction experiment.\label{tab:hyper_param_mix}}
\end{table}
\begin{table}[H]
    \centering
    \begin{tabular}{c|cccc}
    \toprule
    Model & LR & Batch Size (data) & Batch Size (images) & Total Iterations \\
    \midrule
    HOTET (one-to-one) & $10^{-3}$ & 1024 & N/A & 5000\\
    HOTET (multi-to-one) & $10^{-3}$ & 1024 & 8 & 5000\\
    \bottomrule
    \end{tabular}
    \caption{Hyperparamter in the color transfer experiment.\label{tab:hyper_param_color}}
\end{table}

\end{document}